%% file: main.tex
\documentclass{article} % For LaTeX2e
\usepackage{iclr2025_conference,times}

\input{math_commands.tex}

\usepackage{hyperref}
\usepackage{url}

\usepackage{microtype}
\usepackage{graphicx}

\usepackage{subcaption}
\usepackage{booktabs} % for professional tables
\usepackage{xcolor}
\definecolor{darkgreen}{rgb}{0.0, 0.5, 0.0}
\definecolor{darkred}{rgb}{0.6, 0.0, 0.0}
\definecolor{gray}{rgb}{0.5, 0.5, 0.5}
\definecolor{myblue}{RGB}{0, 0, 180}
\definecolor{mygreen}{RGB}{0, 128, 0}
\definecolor{myred}{RGB}{196, 0, 0}

\usepackage{multirow}

\usepackage{hyperref}
\usepackage{url}
\hypersetup{
    colorlinks=true,       % Enable colored links
    linkcolor=myred,      % Color for internal links (sections, pages, etc.)
    citecolor=myblue,     % Color for citation links
    filecolor=magenta,     % Color for file links
    urlcolor=mygreen         % Color for URLs
}

\usepackage{amsmath}
\usepackage{amssymb}
\usepackage{mathtools}
\usepackage{amsthm}

\usepackage{algpseudocode}
\usepackage{algorithm}

\usepackage{enumitem}
\setlist[itemize]{leftmargin=*}

\input{notation}
\input{acronyms}

\title{MaxCutPool: differentiable feature-aware Maxcut for pooling in graph neural networks}

\author{Carlo Abate \thanks{Equal contribution} \\
Alma Mater Studiorum - University of Bologna\\
Fondazione Istituto Italiano di Tecnologia\\
\texttt{carlo.abate@iit.it} \\
\And
Filippo Maria Bianchi $^*$\\
UiT the Arctic University of Norway\\
NORCE Norwegian Research Centre AS \\
\texttt{filippo.m.bianchi@uit.no} \\
}

\iclrfinalcopy % Uncomment for camera-ready version, but NOT for submission.
\begin{document}

\maketitle

\begin{abstract}
We propose a novel approach to compute the \maxcut in attributed graphs, \textit{i.e.}, graphs with features associated with nodes and edges. Our approach works well on any kind of graph topology and can find solutions that jointly optimize the \maxcut along with other objectives.
Based on the obtained \maxcut partition, we implement a hierarchical graph pooling layer for Graph Neural Networks, which is sparse, trainable end-to-end, and particularly suitable for downstream tasks on heterophilic graphs.
\end{abstract}

%%%%%%%%%%%%%%%%%%%%%%%%%%%%%%%%%%%%%
%% INTRODUCTION
%%%%%%%%%%%%%%%%%%%%%%%%%%%%%%%%%%%%%
\section{Introduction}

%% MAXCUT
The \maxcut is the problem of partitioning the nodes of a graph such that as many edges as possible connect nodes from different sides of the partition. 
The \maxcut is orthogonal to the more commonly encountered \mincut, which aims at partitioning the nodes into strongly connected groups.
While \mincut is closely related to clustering, \maxcut relates to the concept of downsampling, \emph{e.g.}, keeping one-every-$K$, under the assumption that there is a redundancy among the $K$ samples.
Like the \mincut, the \maxcut is a combinatorial optimization problem that, in practice, is approximated by approaches that find suboptimal or unstable solutions for a large class of graphs~\citep{makarychev2014bilu}.

%% POOLING GENERAL
Pooling is ubiquitously used in deep learning for gradually reducing the size of the data while retaining important information. In \glspl{cnn}, pooling is typically implemented by selecting the maximum within a contiguous patch (\textit{max-pool}) or by computing an average (\textit{avg-pool}). These strategies are naturally related to \maxcut and \mincut problems, respectively.
%% POOLING IN GNNS
Similarly to \glspl{cnn}, \glspl{gnn}, which can be seen as a generalization to irregular data, are typically built by alternating \gls{mp} and graph pooling layers \citep{zhou2020review}. A hierarchy of pooling layers gradually extracts global graph properties through the computation of local summaries and is key to building deep \glspl{gnn} for graph classification~\citep{Khasahmadi2020Memory-Based}, node classification~\citep{gao2019graph, ma2020path}, graph matching~\citep{liu2021hierarchical}, and spatio-temporal forcasting~\citep{cini2024graph, marisca2024graph}.

%% SOTA POOLING
Two important approaches are followed when implementing hierarchical graph pooling. 
One is to account for the node features with trainable functions that are adapted to a downstream task at hand. The other is to optimize graph theoretical objectives, such as the \mincut or the \maxcut, to guide the computation of the coarsened graph.
Combining the first approach with \mincut objectives is relatively straightforward, as they complement the smoothing effect of \gls{mp} layers~\citep{hansen2023total}.
Conversely, objectives such as \maxcut that select sparse and uniformly distributed subsamples of nodes have been implemented so far only within non-differentiable frameworks, which account neither for node features nor for task objectives~\citep{luzhnica2019clique}.

%% CONTRIBUTIONS
\subsection{Contributions}

\paragraph{\maxcut for attributed graphs.}
Our first contribution is graph theoretical and consists of a novel \gls{gnn}-based approach to compute a \maxcut partition in attributed graphs. Being differentiable, our method can be seamlessly integrated into a deep-learning framework where other loss functions can influence the \maxcut solution.
Remarkably, our method is also more robust than traditional approaches in computing the \maxcut on non-attributed graphs, as it finds a better cut on most graph topologies.
This makes our contribution relevant to \emph{every} application of the \maxcut problem, such as quantum computing~\citep{zhou2020quantum}, circuit design~\citep{9204635}, statistical physics~\citep{borgs2012convergent}, material science~\citep{liers2004computing}, computer vision~\citep{abbas2022rama}, and quantitative finance~\citep{lee2023quantumized}.

\paragraph{Graph pooling and coarsening.}
The \maxcut application we focus on is the problem of learning a coarsened graph within a \gls{gnn}. In particular, we design a new hierarchical pooling layer that reduces the graph by keeping the nodes from one side of the \maxcut partition. Our layer is the first to combine a graph theoretical \maxcut objective with a pooling approach that is features-aware and trainable end-to-end. When we include the newly proposed pooling layer in \glspl{gnn} for graph and node classification, we achieve similar or superior performances compared to state-of-the-art pooling techniques.

\paragraph{Improved scoring-based pooling framework.}
We propose a simple and efficient scheme to assign nodes to supernodes when computing the pooled graph.
Our scheme can be applied not only to our method but to the whole family of sparse scoring-based graph pooling operators enhancing, in principle, their representational power.
Importantly, we bridge the gap between scoring-based and dense pooling methods by using the same operations to compute the features and the topology of the pooled graph.

\paragraph{Heterophilic graph classification dataset.}
Differently from the existing differentiable pooling operators, the nature of the \maxcut solution makes our graph pooling operator particularly suitable for heterophilic graphs.
While there are benchmark datasets for node classification on heterophilic graphs, there is a lack of such datasets for graph classification.
To fill this gap, we introduce a novel synthetic dataset that, to our knowledge, is the first of its kind.

%%%%%%%%%%%%%%%%%%%%%%%%%%%%%%%%%%%%%
%% BACKGROUND
%%%%%%%%%%%%%%%%%%%%%%%%%%%%%%%%%%%%%
\section{Background}

\subsection{The \maxcut problem and the continuous relaxations} 
\label{sec:maxcut}

Let $\gG = (\gV,\gE)$ be an undirected graph with non-negative weights on the edges, and let $N$ be the number of nodes in $\gG$. A cut in $\gG$ is a partition $(\gS, \gV \setminus \gS)$ where $\gS \subset \gV$. The \maxcut problem consists of finding a cut that maximizes the total volume of edges connecting nodes in $\gS$ with those in $\gV \setminus \gS$. 
The \maxcut objective can be expressed as the integer quadratic problem
\begin{equation}
    \label{eq:maxcut}
    \max_{\vz} \sum \limits_{i,j \in  \mathcal{V}} w_{ij}(1- z_i z_j) \;\; \text{s.t.} \;\; z_i \in \{ -1, 1 \},
\end{equation}
where $\vz \in \{-1,1\}^N$ is an assignment vector indicating to which side of the partition each node is assigned and $w_{ij}$ is the weight of the edge connecting nodes $i$ and $j$.

Like other discrete optimization problems of this kind, \maxcut is NP-hard.
The \gls{gw} algorithm \citep{goemanswilliamson1995improved} provides a semidefinite relaxation of the integer quadratic problem, which makes it tractable:
\begin{equation}
    \label{eq:maxcut_relaxation}
    \max_{\mX} \sum \limits_{i,j \in  \mathcal{V}} w_{ij}(1 - \vx_i \cdot \vx_j) \;\; \text{s.t.} \;\; \Vert \vx_i \Vert = 1,
\end{equation}
where $\mX \in \sR^{N \times D}$ is a matrix whose rows are the continuous embeddings of the nodes in $\gG$.
The vectors $\mX$ are projected on a random hyperplane to split the nodes and assign them to the two sides of the partition. This algorithm guarantees an expected cut size of $.868$ of the maximum cut.

Another simple yet effective continuous relaxation is the \gls{levs} method~\citep{shuman2015multiscale}.
Let $\mL$ be the Laplacian matrix associated to the graph $\gG$ and let $\vu_{\max}$ be the eigenvector of $\mL$ associated to the largest eigenvalue $\lambda_{\max}$. A cut in $\gG$ can be found based on the polarity of the components of $\vu_{\max}$, for instance by letting $\gS = \{i: \vu_{\max}[i] \geq 0 \}$.
In the field of graph signal processing, the eigenvectors related to the largest eigenvalues of $\mL$ are closely related to the operation of high-pass filtering of a graph signal~\citep{tremblay2018design}. Specifically, they are used to design graph filters that amplify \textit{high-frequency} components of a signal, \emph{i.e.}, the components that vary the most across adjacent nodes~\citep{shuman2013emerging}.

The \maxcut problem is closely related to \textit{graph coloring}, which aims at assigning different colors to adjacent nodes.
In particular, the 2-color \emph{approximate coloring}~\citep{1984Graph}, is the problem of identifying subsets of nodes such that the connections within each subset are minimized. Such coloring is a high-frequency graph signal and induces a partition that is orthogonal with respect to spectral clustering~\citep{vonluxburg2007tutorial}.

\begin{figure}[htbp]
    \centering
    \begin{subfigure}[b]{0.24\textwidth}
        \centering
        \includegraphics[width=.95\textwidth]{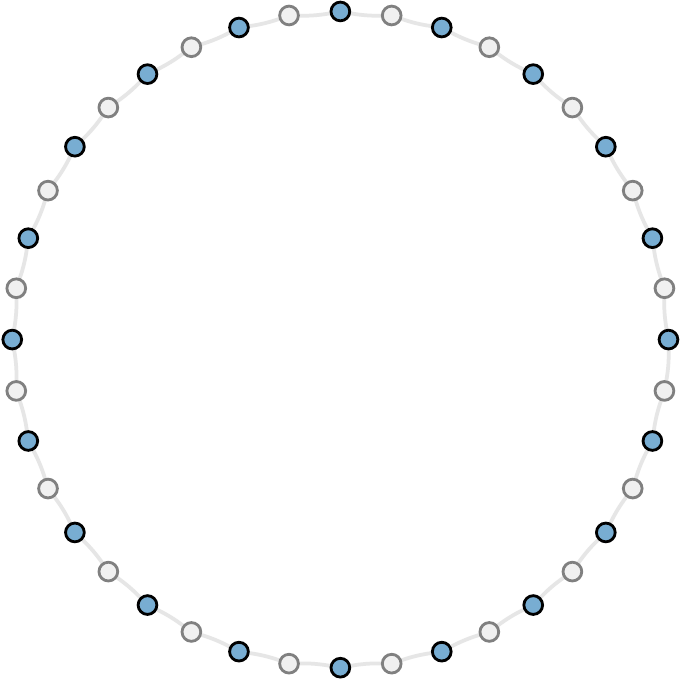}
        \caption{}
    \end{subfigure}
    \hfill
    \begin{subfigure}[b]{0.24\textwidth}
        \centering
        \includegraphics[width=.9\textwidth]{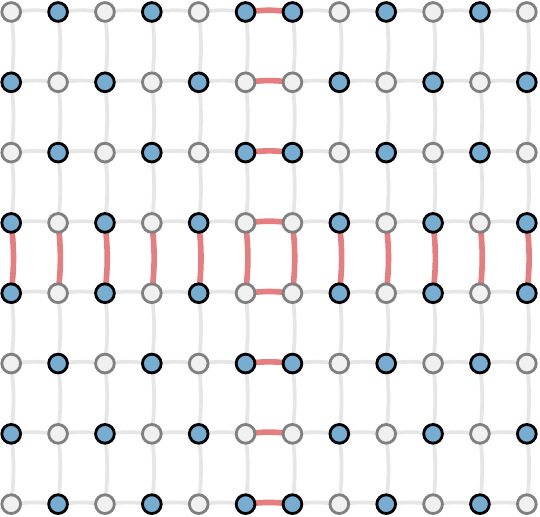}
        \caption{}
    \end{subfigure}
    \hfill
    \begin{subfigure}[b]{0.24\textwidth}
        \centering
        \includegraphics[width=.95\textwidth]{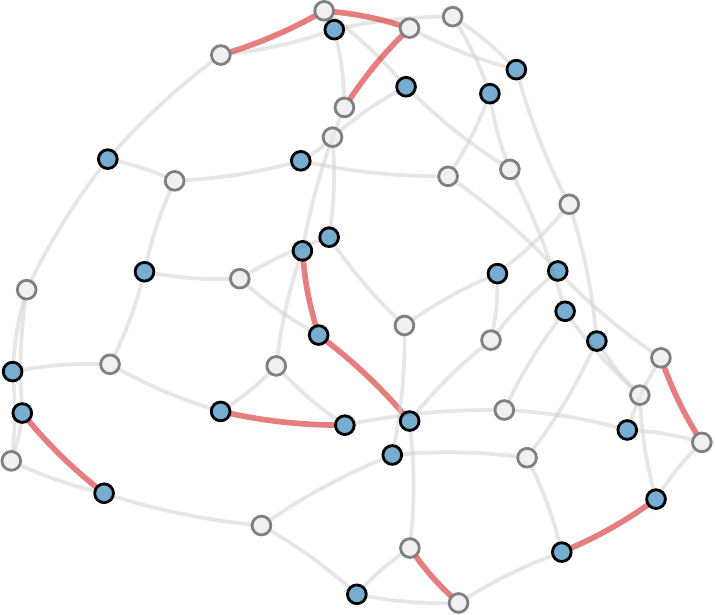}
        \caption{}
    \end{subfigure}
    \hfill
    \begin{subfigure}[b]{0.24\textwidth}
        \centering
        \includegraphics[width=.9\textwidth]{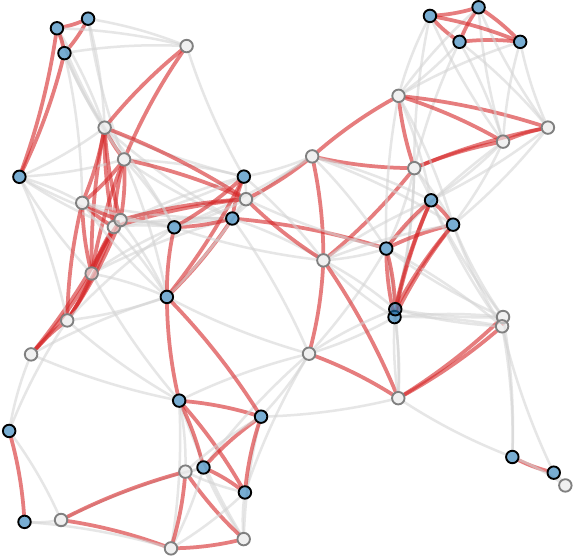}
        \caption{}
    \end{subfigure}

    \begin{subfigure}[b]{0.24\textwidth}
        \centering
        \includegraphics[width=\textwidth]{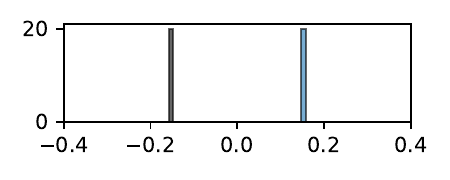}
        \caption{}
    \end{subfigure}
    \hfill
    \begin{subfigure}[b]{0.24\textwidth}
        \centering
        \includegraphics[width=\textwidth]{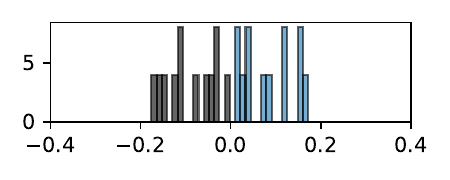}
        \caption{}
    \end{subfigure}
    \hfill
    \begin{subfigure}[b]{0.24\textwidth}
        \centering
        \includegraphics[width=\textwidth]{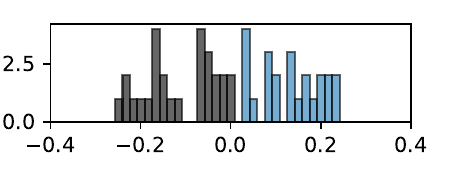}
        \caption{}
    \end{subfigure}
    \hfill
    \begin{subfigure}[b]{0.24\textwidth}
        \centering
        \includegraphics[width=\textwidth]{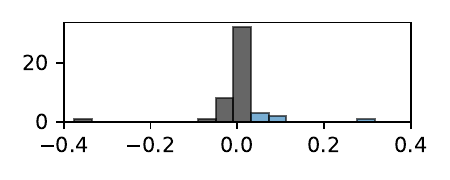}
        \caption{}
    \end{subfigure}

    \begin{subfigure}[b]{0.24\textwidth}
        \centering
        \includegraphics[width=\textwidth]{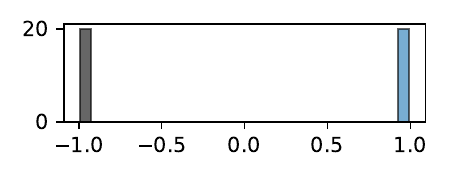}
        \caption{}
    \end{subfigure}
    \hfill
    \begin{subfigure}[b]{0.24\textwidth}
        \centering
        \includegraphics[width=\textwidth]{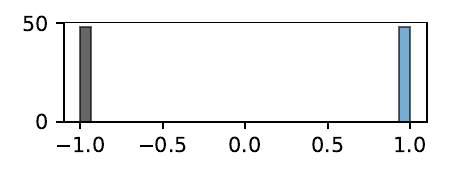}
        \caption{}
    \end{subfigure}
    \hfill
    \begin{subfigure}[b]{0.24\textwidth}
        \centering
        \includegraphics[width=\textwidth]{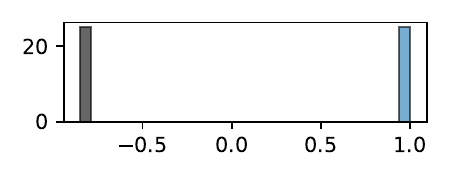}
        \caption{}
    \end{subfigure}
    \hfill
    \begin{subfigure}[b]{0.24\textwidth}
        \centering
        \includegraphics[width=\textwidth]{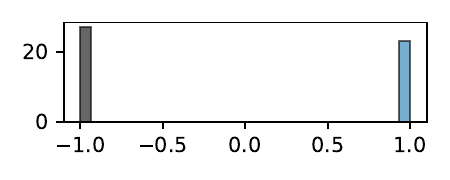}
        \caption{}
    \end{subfigure}

    \caption{\textbf{Top row:} Partitions induced by the sign of the elements in $\vu_{\max}$. The nodes are colored based on the partition and the red edges are those not cut (the less, the better). \textbf{Middle row:} histograms of $\vu_{\max}$ inducing the partitions above. While in bipartite graphs the separation is sharp, the more a graph is irregular and dense the more the values are clustered around zero, making it difficult to find the optimal \maxcut.
    \textbf{Bottom row:} histograms of the score vectors generated by our model, which always produce a clear and sharp partition.
    }
    \label{fig:vmax}
\end{figure}

A \maxcut partition that cuts every edge exists only for bipartite graphs. 
Conversely, in fully connected graphs no more than half of the edges can be cut.
Algorithms relying on continuous relaxations to find the \maxcut partition tend to be unstable and perform poorly, especially when the graph topology departs from the bipartite case~\citep{trevisan2009max}.
Fig.~\ref{fig:vmax}(a-h) shows the performance of the \gls{levs} method on bipartite and non-bipartite graphs: numerical issues typically occur in more dense and less regular graphs, making it difficult to identify the optimal \maxcut solution.

\subsection{Message passing in GNNs}
Let us consider a graph $\gG = \left(\mX \in \sR^{N \times F}, \mA \in \sR^{N \times N} \right)$. 
A basic \gls{mp} operator can be described as
\begin{equation}
    \label{eq:mp}
    \mX' = \sigma \left( \mP \mX \boldsymbol{\Theta} \right)
\end{equation}
where $\sigma$ is a non linear activation function, $\boldsymbol{\Theta}$ are trainable parameters, and $\mP$ is a propagation operator matching the sparsity pattern of $\mA$. 
Each \gls{mp} layer relies on a specific propagation operator. 
For instance, in \glspl{gcn}~\citep{kipf2017semisupervised}, the propagation operator is defined as $\mP = \Hat{\mD}^{-\frac{1}{2}}\Hat{\mA}\Hat{\mD}^{-\frac{1}{2}}$, where $\Hat{\mA} = \mA + \mI$ and $\Hat{D}_{ii} = \sum_{j=0}\Hat{\mA}_{ij}$.

Due to the fixed, non-negative smoothing nature of common propagation operators, the repeated application of $\mP$ can lead to over-smoothing. 
If that happens, the feature representations of nodes become increasingly similar, hindering the function approximation capabilities of a \gls{gnn}, which can only learn smooth graph signals~\citep{wu2019oversmoothing, wang2019oversmoothing}. 
In contrast, by combining smoothing propagation operators with \emph{sharpening} ones, any kind of gradients can be learned~\citep{eliasof2023improving, bianchi2021graph}.
Since there is no universal definition for such an operator, we rely on the formulation introduced by \cite{bianchi2022simplifying}:
\begin{equation}
\label{eq:het_mp}
    \mP = \mI - \delta \left( \mI - \mD^{\frac{1}{2}}\mA\mD^{\frac{1}{2}} \right) = \mI - \delta\mL^{\text{sym}}
\end{equation}
where $\delta$ is a smoothness hyperparameter and $\mL^{\text{sym}}$ is the symmetrically normalized Laplacian of $\gG$. 
As observed by \cite{bianchi2022simplifying}, when $\delta = 0$ the \gls{mp} behaves like a simple \gls{mlp}. Instead, when $\delta = 1$ the behavior is close to that of a \gls{gcn}. 
Finally, as noted by \cite{eliasof2023improving}, when $\delta > 1$ the propagation operator favors the realization of non-smooth signals on the graph and we refer to this variant as a \gls{hetmp} operator. We note that this can be seen as the graph counterpart of the Laplacian sharpening kernels for images, mapping connected nodes to different values~\citep{mather2022computer}.

\subsection{Graph pooling}
While there are profound differences between existing graph pooling approaches, most of them can be expressed through the \gls{src} framework \citep{grattarola2022understanding}. 
Specifically, a pooling operator $\texttt{POOL}: (\mA,\mX) \mapsto (\mA', \mX')$ can be expressed as the combination of three sub-operators:

\begin{itemize}
    \item $\texttt{SEL}:(\mA,\mX) \mapsto \mS \in \sR^{N \times K}$, is a selection operator that defines how the $N$ original nodes are mapped to the $K$ pooled nodes, called \emph{supernodes}, being $\mS$ the \emph{selection matrix}. 
    \item $\texttt{RED}:(\mX, \mS) \mapsto \mX' \in \sR^{K \times F}$, is a reduction operator that yields the features of the supernodes. A common way to implement \texttt{RED} is $\mX' = \mS^\top\mX$.
    \item $\texttt{CON}:(\mA, \mS) \mapsto \mA' \in \sR^{K \times K}_{\geqzero}$, is a connection operator that generates the new adjacency matrix and, potentially, edge features. Typically, \texttt{CON} is implemented as $\mA' = \mS^\top\mA\mS$ or $\mA' = \mS^{+}\mA\mS$.
\end{itemize}

Different design choices for \texttt{SEL}, \texttt{RED}, and \texttt{CON} induce a taxonomy of the operators. 
For example, if \texttt{SEL}, \texttt{RED}, and \texttt{CON} are learned end-to-end the pooling operators are called \emph{trainable}, \emph{non-trainable} otherwise. 
Relevant to this work, are the families of pooling methods described in the following.

% SOFT CLUSTERING
\textbf{Soft-clustering} methods, sometimes referred to as \emph{dense}~\citep{grattarola2022understanding}, assign each node to more than one supernode through a soft membership. 
Representatives methods such as DiffPool~\citep{ying2018hierarchical}, \acrlong{mincut}~\citep{bianchi2020spectral}, 
StructPool~\citep{yuan2020structpool},
HoscPool~\citep{duval2022higher}, 
and \gls{dmon}~\citep{tsitsulin2020graph}, compute a soft cluster assignment matrix $\mS \in \R^{N \times K}$ either with an \gls{mlp} or an \gls{mp}-layer operating on node the features and followed by a \texttt{softmax}. 
Each method leverages different unsupervised auxiliary loss functions to guide the formation of the clusters. 
Trainable soft-clustering methods usually perform well on downstream tasks due to their flexibility and \emph{expressive power}, which is the capability of retaining all the information from the original graph~\citep{bianchi2023expr}.
However, storing the soft assignments $\mS$ is a memory bottleneck for large graphs (see, \emph{e.g.}, the analysis of memory usage in Appendix~\ref{app:memory_usage}) and soft memberships cause pooled graphs to be very dense and not interpretable. 
Additionally, each graph is mapped to the same fixed number of supernodes $K$, which can hinder the generalization capabilities in datasets where the size of each graph varies significantly.

\textbf{Scoring-based} methods select supernodes from the original nodes based on a node scoring vector $\vs$. 
The chosen nodes correspond to the top $K$ elements of $\vs$, where $K$ can be a ratio of the nodes in each graph, making these methods adaptive to the graph size. 
Representatives such as \gls{topk}~\citep{gao2019graph, knyazev2019understanding}, ASAPool~\citep{ranjan2020asap}, SAGPool~\citep{lee2019self}, PanPool~\citep{ma2020path}, TAPool~\citep{gao2021topologytapool}, CGIPool~\citep{pang2021graph}, and IPool~\citep{gao2022ipool} primarily differ in how they compute the scores or in the auxiliary tasks they optimize to improve the quality of the pooled graph. 
Despite a few attempts to encourage diversity among the selected nodes~\citep{zhang2019hierarchical, noutahi2019towards}, scoring-based methods derive the scores from node features that tend to be locally similar, especially after being transformed by \gls{mp} operations.
As such, the pooled graph often consists of a chunk of strongly connected nodes that possess similar characteristics. 
Consequently, entire sections of the graph are usually not represented, leading to reduced expressiveness and lower performance in downstream tasks~\citep{wang2024comprehensive}.

\textbf{One-every-$\boldsymbol{K}$} methods leverage graph-theoretical properties to select supernodes by subsampling the graph uniformly. 
For instance, \gls{kmis} \citep{bacciu2023pooling} identifies as supernodes the members of a maximal $K$-independent set, \emph{i.e.}, nodes separated by at least $K$-hops on the graph. 
Graclus~\citep{dhillon2007weighted, defferrard2016convolutional} creates supernodes by merging the pairs of most connected nodes in the graph.
SEP~\citep{wu2022structural} partitions the node hierarchically according to a precomputed tree that minimizes the structural entropy of the graph.
\gls{ndp}~\citep{bianchi2020hierarchical} divides the graph into two sets, $\gV_+$ and $\gV_-$, according to the partition induced by the components of $\vu_{\max}$ (see Section \ref{sec:maxcut}).
One of the two sides of the partition is dropped ($\gV_-$), while the other ($\gV_+$) becomes the set of supernodes.
While both Graclus and \gls{ndp} can only reduce the number of nodes by approximately half, higher pooling ratios (one-every-$2^K$) are achieved by applying them recursively $K$ times.
Nevertheless, they lack the same control of soft-clustering and scoring-based methods in fixing the size of the pooled graphs.
Like the scoring-based methods, one-every-$K$ methods are adaptive and produce crisp cluster assignments.
However, they are not trainable and precompute the pooled graph based on the topology without accounting for the node features or the downstream task. 
Tab.~\ref{tab:drawbacks} summarizes the drawbacks of the existing families of pooling methods. 

\input{tables_2025/pooling_drawbacks}

%%%%%%%%%%%%%%%%%%%%%%%%%%%%%%%%%%%%%
%% METHOD
%%%%%%%%%%%%%%%%%%%%%%%%%%%%%%%%%%%%%
\section{Method}
\label{sec:method}

We leverage a GNN to generate a \maxcut partition while accounting for node features and additional objectives from a downstream task. 
In particular, we let node features and task-specific losses influence the selection of the \maxcut solution, creating partitions that not only maximize the number of cut edges but also prioritize nodes that are optimal for the downstream task.
To reach this goal, it is necessary to overcome a tension between the effect of a standard MP layer and the \maxcut: the first applies a smoothing operation that makes adjacent nodes similar, which is orthogonal to the objective of the latter. 
Therefore, to implement \maxcut with a \gls{gnn} we rely on \gls{hetmp}, implemented by setting $\delta > 1$ in the \gls{mp} operation in Eq.~\ref{eq:het_mp}. 
As discussed in Sec. \ref{sec:maxcut}, solving the \maxcut problem is equivalent to coloring adjacent nodes differently. Notably, this is an intrinsic effect of \gls{hetmp} that makes features of adjacent nodes as different as possible, effectively acting as a high-pass graph filter. 
Therefore, optimizing a \maxcut loss on features generated by \gls{hetmp} layers overcomes the limitation of traditional scoring-based methods that compute the scores from features produced by homogeneous MP operators.

The layer we propose is called \gls{diffndp} and we present it through the \gls{src} framework. 
The \texttt{SEL} operation in \gls{diffndp} identifies as supernodes a subset $\gS$ of the nodes in the original graph. An auxiliary GNN, called \emph{ScoreNet}, consists of a stack of \gls{hetmp} layers that map the node features into a vector $\boldsymbol{s} = \text{ScoreNet}(\mX,\mA) \in [-1,1]^N$, which assigns a score to each node.  
The indices $\vi = \text{top}_K(\vs)$ associated with the highest scores identify the $K$ supernodes.
Additional details about the ScoreNet are in Appendix~\ref{app:auxnet}.
Fig.~\ref{fig:vmax}(i-l) shows the histograms of the score vectors $\boldsymbol{s}$ generated by the ScoreNet for the $4$ example graphs.
Compared to the histograms of $\vu_\text{max}$ in Fig.~\ref{fig:vmax}(e-h), the values in $\boldsymbol{s}$ always produce a distribution with two sharp and well-separated modes, yielding a clear node partition.

\begin{figure}[htbp]
    \centering
    \begin{subfigure}[b]{0.19\textwidth}
        \centering
        \includegraphics[width=\textwidth]{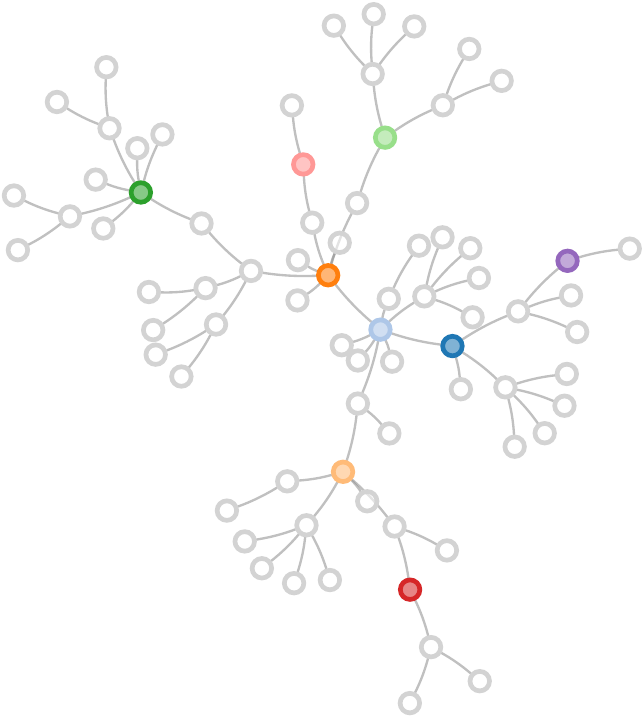}
        \caption{}
    \end{subfigure}
    \hfill
    \begin{subfigure}[b]{0.19\textwidth}
        \centering
        \includegraphics[width=\textwidth]{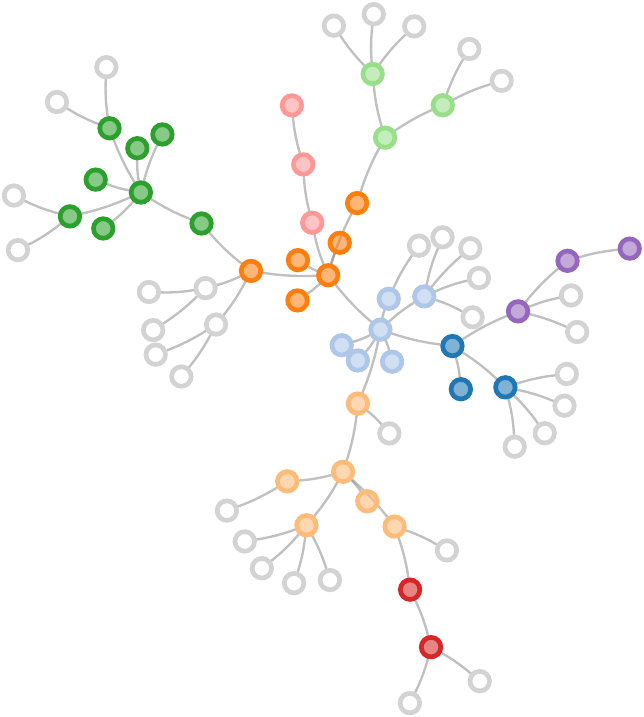}
        \caption{}
    \end{subfigure}
    \hfill
    \begin{subfigure}[b]{0.19\textwidth}
        \centering
        \includegraphics[width=\textwidth]{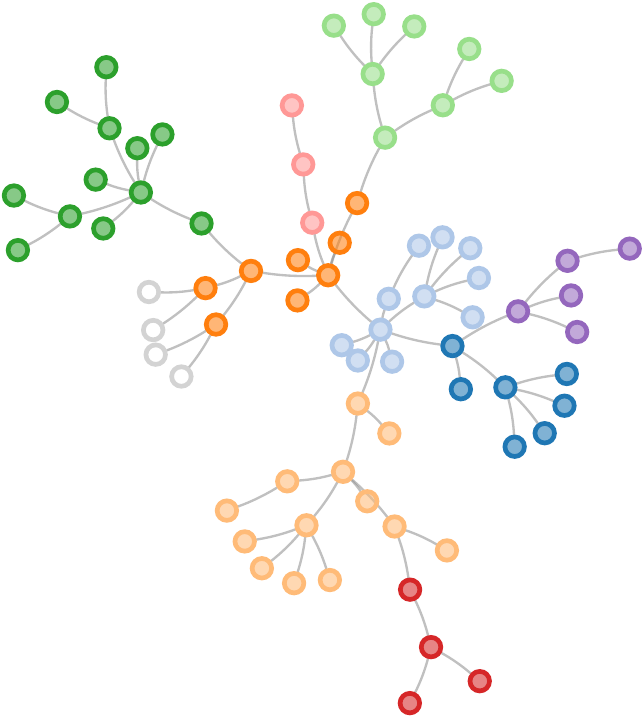}
        \caption{}
    \end{subfigure}
    \hfill
    \begin{subfigure}[b]{0.19\textwidth}
        \centering
        \includegraphics[width=\textwidth]{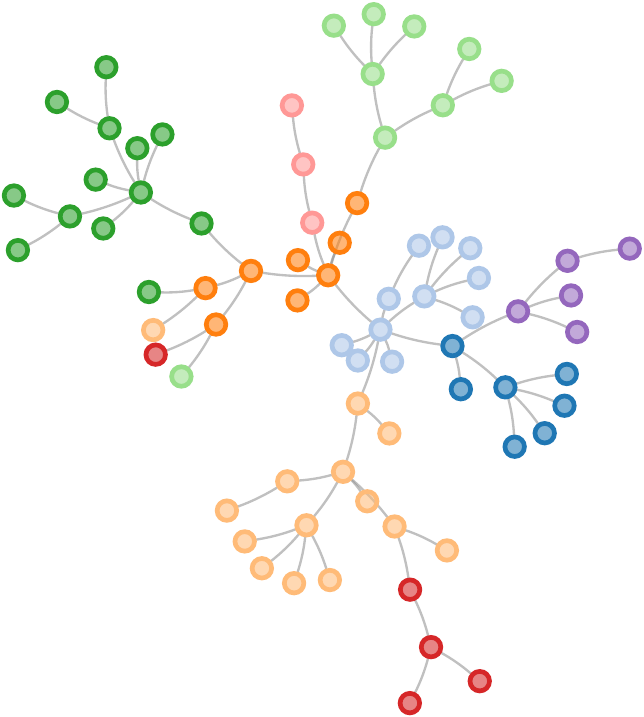}
        \caption{}
    \end{subfigure}
    \hfill
    \begin{subfigure}[b]{0.19\textwidth}
        \centering
        \includegraphics[width=\textwidth]{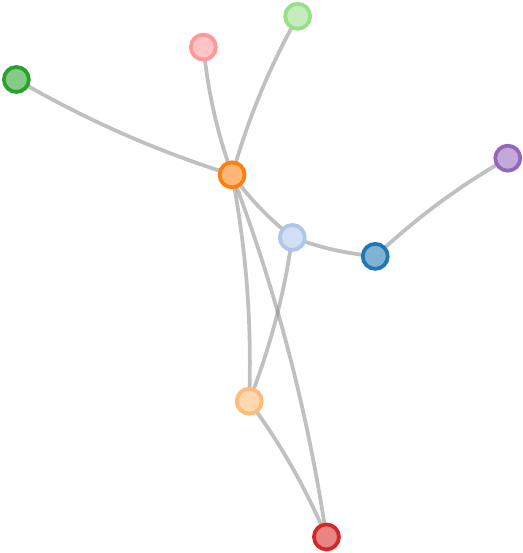}
        \caption{}
    \end{subfigure}
    
    \caption{\textbf{(a)} The nodes with the $K=9$ highest scores are selected. \textbf{(b-c)} Their ID is propagated to the unselected nodes until all are covered or until a maximum number of iterations ($2$ here) is reached. \textbf{(d)} The $4$ remaining nodes are assigned randomly. \textbf{(e)} The pooled graph is obtained by aggregating the nodes with the same ID and coalescing the edges connecting nodes from different groups.}
    \label{fig:propagation}
\end{figure}

After the $K$ supernodes are selected, the remaining $N-K$ nodes are assigned to one of the supernodes via the nearest neighbor aggregation.
An assignment matrix $\mS$ is built by performing a breadth-first visit of the graph where, starting from the supernodes, all the remaining nodes are assigned to their nearest supernode (see Figure \ref{fig:propagation}).
More formally, the assignment matrix $\mS$ is defined as
$$\texttt{SEL}: [\mS]_{ij} = 1 \iff j = \phi(\gS, \mA,i),$$ 
where $\phi(\gS, \mA, i)$ returns the closest supernode to node $i$. 
The visit of the graph is iterated until all nodes are assigned to a supernode or until a maximum number of iterations is reached.
In the latter case, nodes that are still unassigned are assigned at random to ensure that the pooled graph is always connected.
Keeping the maximum number of iterations small (\emph{e.g.}, $2$ or $3$) prevents pointless attempts to reach a supernode, \emph{e.g.}, when there are no supernodes within a disconnected graph component, and also injects randomness acting as a regularizer that helps to move away from suboptimal configurations often encountered at the beginning of the training stage.
The pseudo-code for a GPU-parallel implementation of the proposed assignment scheme is deferred to Appendix~\ref{app:nearest-neighbour}. 

The \texttt{RED} operation for computing the features of the supernodes can be implemented in two ways: 
\begin{center}
    \texttt{RED}: $\mX' = s_\vi \odot [\mX]_{\vi}$ \;\; or \;\;
    \texttt{RED}: $\mX' = \vs_{\vi} \odot 
    \mS^\top\mX$.
\end{center}
The Hadamard product $\odot$ enables gradients flowing through the ScoreNet during back-propagation, making it possible for the gradients of the task loss to reach every component of our model, despite the non-differentiable top$_K$ operation.
When the first or the second variant is used to implement \texttt{RED}, we refer to the pooling operator as \gls{diffndp} and \gls{diffndp}-E, respectively.
The suffix ``-E'' indicates that the pooling operator satisfies the sufficient conditions for expressiveness defined by \cite{bianchi2023expr}.

Finally, the \texttt{CON} operation can be implemented as 
$$\texttt{CON}: \mA' = \mS^\top\mA\mS.$$

%%%% AUX LOSS %%%%
\subsection{Auxiliary loss}

Each \gls{diffndp} layer is associated with an auxiliary loss that encourages the top-$K$ selected nodes to belong to the same side of the \maxcut partition.
The loss is defined as:
\begin{equation} 
\label{eq:loss}
    \mathcal{L}_{\text{cut}} = \frac{\vs^\top A \vs}{|E|}
\end{equation}
where $|E| = \sum_{ij} w_{ij}$ is the total edge weight of the graph. Since $-1 \leq s_i \leq 1$, we have $-|E| \leq \vs^\top A \vs \leq |E|$, hence $-1 \leq \mathcal{L}_{\text{cut}} \leq 1$. 
This loss evaluates the ratio between the volume of the cut and the total volume of the edges. 
Minimizing $\mathcal{L}_{\text{cut}}$ encourages the nodes to be assigned to different partitions if and only if they are connected. 
The loss reaches its minimum $-1$ when all connected nodes are assigned to opposite sides of the partition, \emph{i.e.}, when all the edges are cut.
Clearly, this can happen only in bipartite graphs. The details about the derivation of the loss are in Appendix~\ref{app:aux_loss}.

A GNN model consisting of \gls{mp} layers interleaved with \gls{diffndp} layers can be trained end-to-end to jointly minimize a task loss $\mathcal{L}_{\text{task}}$ and the auxiliary loss $\mathcal{L}_{\text{cut}}$.
The total loss is then defined as
\begin{equation}
    \mathcal{L} = \mathcal{L}_{\text{task}} + \sum_l \beta \mathcal{L}^{(l)}_{\text{cut}}
\end{equation}
where $\beta$ is a scalar weighting each auxiliary loss $\mathcal{L}^{(l)}_{\text{cut}}$ associated with the $l$-th \gls{diffndp} layer.

\subsection{Homophilic and heterophilic operations}
Despite the presence of \gls{hetmp} layers and the heterophilic loss, \gls{diffndp} can be inserted into \glspl{gnn} equipped with traditional \gls{mp} layers. In fact, the homophilic nature of the latter is leveraged by our method. After a standard homophilic \gls{mp}, the stronger the association between a pair of nodes, the more similar their features will be.
Keeping them both will, thus, be redundant, and one of them can be dropped.
This is precisely what is done by \texttt{SEL} that, thanks to its heterophilic design, samples nodes that share as few connections as possible and are uniformly distributed over the graph. 

\gls{diffndp} contains an additional homophilic operation: the computation of the assignments $\mS$ through the nearest-neighbor aggregation, which synergizes with the uniform sampling of the supernodes.
The assignments are used by \texttt{CON} to produce a connectivity matrix that is connected yet sparse
and they can also be leveraged by \texttt{RED} to ensure the expressiveness of the pooling layer.
However, in heterophilic datasets where such a homophilic assignment might not be appropriate, the non-expressive variant of our method offers a more suitable alternative.

In general, rather than creating tension, combining homophilic and heterophilic components provides a \gls{gnn} with \gls{diffndp} the flexibility for handling different scenarios. As we show in the experimental evaluation, our model can even switch to a completely homophilic setting by adjusting the values of $\delta$ and $\beta$ or by ignoring the auxiliary loss $\mathcal{L}_{\text{cut}}$.

\subsection{Relation with other pooling methods}

\gls{diffndp} belongs to the scoring-based pooling family from which it inherits the possibility of specifying any desired pooling ratio adaptable to the size $N_i$ of the $i$-th graph, \emph{i.e.}, $K_i = \floor{N_i * 0.5}$ while achieving node selection patterns similar to one-every-$K$ methods.
In methods like \gls{kmis} the flexibility provided by trainable functions is only used to choose between a small set of maximal solutions that do not break the hard constraint of the supernodes to be $K$-independent.
On the other hand, in \gls{diffndp} any set of supernodes can, in principle, be chosen making \gls{diffndp} more flexible and able to better adapt to the requirements of the downstream task. 
Finally, \gls{diffndp} is the only scoring-based pooler with a graph-theoretical auxiliary regularization loss. 

The proposed method for constructing the cluster assignment matrix $\mS$ from the selected supernodes can be applied to other scoring-based pooling methods. 
This naturally enhances their expressiveness by enabling the use of the same \texttt{CON} and \texttt{RED} operations adopted by soft-clustering approaches, which retains all the information from the original graph.
By addressing the key limitation in the expressiveness of scoring-based pooling methods, our approach retains the benefits of sparsity and interpretability of scoring-based poolers while narrowing the gap with soft-clustering methods.

%%%%%%%%%%%%%%%%%%%
%%% EXPERIMENTS %%%
%%%%%%%%%%%%%%%%%%%
\section{Experimental evaluation}
\label{sec:experiments}

We consider three different tasks to demonstrate the effectiveness of \gls{diffndp}.
The code to reproduce the reported results is publicly available\footnote{
%\url{https://anonymous.4open.science/r/MaxCutPool}
\url{https://github.com/NGMLGroup/MaxCutPool}
}.
The details of the architectures used in each experiment and the hyperparameter selection procedure are described in Appendix~\ref{app:architectures}.

\subsection{Computation of the \maxcut partition}
\label{sec:maxcut_experiment}

The main focus of this experiment is to evaluate the capability of the proposed loss $\mathcal{L}_\text{cut}$ to optimize the \maxcut objective, despite the potential risks of getting stuck in local minima due to its gradient-based nature. We compute a \maxcut partition by training a simple \gls{gnn} consisting of a \gls{mp} layer followed by \gls{diffndp} and trained by minimizing only the loss in Eq.~\ref{eq:loss} (details in Appendix~\ref{app:cut_model}).
We use a \gls{gin} layer \citep{xu2018powerful} as the \gls{mp} layer. 
We compare our model against the \gls{levs} approach based on $\vu_\text{max}$, the \gls{gw} algorithm, and a \gls{gnn} with \gls{gcn} layers that minimize a \maxcut loss, as proposed by~\cite{schuetz2022combinatorial}.
For a fair comparison, ours and this \gls{gnn} architecture have a comparable number of learnable parameters.

We considered $9$ graphs generated via the PyGSP library \citep{defferrard2017pygsp}, including bipartite graphs such as the Grid2D and Ring, and $7$ graphs from the GSet dataset~\citep{ye2003gset}, including random, planar, and toroidal graphs, typically used as benchmark for evaluating \maxcut algorithms (details in App.~\ref{app:cut-datasets}). 
Results are shown in Tab.~\ref{tab:maxcut}. 
Performance is computed in terms of the percentage of cut edges: the higher, the better.
With one exception, \gls{diffndp} always finds the best cut.

\input{tables_2025/maxcut_res}

\subsection{Graph classification}

For this task, we evaluate the classification accuracy of a GNN classifier with the following structure: \gls{mp}(32)-Pool-\gls{mp}(32)-Readout, using \gls{gin} as the \gls{mp} layer.
The Pool operation is implemented either by \gls{diffndp} or by the following competing methods:
Diffpool~\citep{ying2018hierarchical}, \gls{dmon}~\citep{tsitsulin2020graph}, 
\acrlong{mincut}~\citep{bianchi2020spectral},
\gls{topk}~\citep{gao2019graph},
Graclus~\citep{dhillon2007weighted}, 
\gls{kmis}~\citep{bacciu2023pooling}.
We also consider \gls{ecpool}~\citep{diehl2019edge} that pools the graph by contracting the edge connecting similar nodes.
%takes an edge-view pooling approach, which is different from the node-view approach of the other methods~\citep{zhou2022edge}.
For \gls{diffndp}, we evaluate three variants:
(i) \gls{diffndp}, the standard version;
(ii) \gls{diffndp}-E, the variant with expressive \texttt{CON};
(iii) \gls{diffndp}-NL, where ``NL'' stands for ``no loss'', meaning we do not optimize the auxiliary loss in the GNN. This serves as an ablation study to assess the importance of the auxiliary loss.
Whenever edge attributes are available, the first \gls{gin} layer is replaced by a GINE layer~\citep{hu2020strategies-gine}, which takes into account edge attributes. Further implementation details are in Appendix \ref{app:graph_class}.

As graph classification datasets we consider $8$ TUD datasets (COLLAB, DD, NCI1, ENZYMES, MUTAG, Mutagenicity, PROTEINS, and REDDIT-BINARY)~\citep{morris2020tudataset}, the \gls{gcb}~\citep{bianchi2022pyramidal}, and EXPWL1~\citep{bianchi2023expr}, which is a recent dataset for testing the expressive power of GNNs. 
In addition, we introduce a novel dataset consisting of $5,000$ multipartite graphs:
each graph is complete $10$-partite, meaning that the nodes can be partitioned into $10$ groups so that the nodes in each group are disconnected, but are connected to all the nodes of the other groups. 
To the best of our knowledge, this is the first benchmark dataset for graph classification with heterophilic graphs.
While the Multipartite dataset consists of complex graph structures, the classification label is determined solely by the node features, allowing us to assess whether the \gls{gnn} can effectively isolate relevant information despite the presence of misleading topological information.
The construction of the Multipartite dataset and a further discussion about its properties are reported in Appendix~\ref{app:multipartite}.

Whenever the node features were not available, we used node labels. 
If node labels were also unavailable, we used a constant as a surrogate node feature. Further details about the remaining datasets can be found in Appendix~\ref{app:graph-classification-datasets}.
The datasets were split via a $10$-fold cross-validation procedure. 
The training dataset was further partitioned into a 90-10\% train-validation random split. 
This approach is similar to the procedure described by \cite{Errica2020A}. 
Each model was trained for $1,000$ epochs with early stopping, keeping the checkpoint with the best validation accuracy.

\input{tables_2025/graph_class_student_fixed}

The results are reported in Tab.~\ref{tab:graph-classification}. 
For completeness, we also reported the performance of the same GNN model without pooling layers (``No pool'').
We conducted a preliminary ANOVA test ($p$-value $0.05$) for each dataset followed by a pairwise Tukey-HSD test ($p$-value $0.05$) to group models whose performance is not significantly different.
Those belonging to the top-performing group are colored in green.
The ANOVA test failed on ENZYMES, PROTEINS, MUTAG, and DD, meaning that the difference in the performance of the GNNs equipped with different poolers is not significant. For this reason, the results on these datasets are omitted from Tab.~\ref{tab:graph-classification} and reported in Appendix~\ref{app:graph_class_extra}.

\gls{diffndp} consistently ranks among the top-performing methods across all evaluated datasets.
Notably, on the EXPWL1 even the non-expressive variant of \gls{diffndp} achieves a perfect accuracy (100\%), outperforming the competitors. This is the first known example of a non-expressive pooler passing the expressiveness test provided by this dataset.
On the Multipartite dataset, \gls{diffndp} performs significantly better than every pooling method. 
When compared to the ``No pool'' baseline, on most datasets \gls{diffndp} improves the classification performance by increasing the receptive field of the \gls{mp} layers while retaining only the necessary information and enhancing the overall expressive power of the \gls{gnn} model.
It is worth noting that the EXPWL1 and Multipartite are the least homophilic datasets (see Appendix~\ref{app:graph-classification-datasets}), indicating that \gls{diffndp} is particularly effective for heterophilic graphs.
On the COLLAB dataset, all \gls{diffndp} variants achieve the top accuracy of $77\%$, showing a statistically significant improvement over other methods. Notably, in the \gls{diffndp} and \gls{diffndp}-E variants the auxiliary loss term plateaued around $0$, making them equivalent to the \gls{diffndp}-NL variant that, in this case, achieves the same performance.
This indicates that our method remains robust even when the auxiliary loss is not needed for the downstream task. 
Overall, the \gls{diffndp}-E variant, which satisfies expressiveness conditions, exhibits similar or better performance compared to \gls{diffndp} across most datasets. In contrast, the performance decline observed in the \gls{diffndp}-NL variant demonstrates the importance of the auxiliary loss.

\subsection{Node classification}
\label{sec:node_class}

For this task, we adopted a simple auto-encoder architecture for node classification: \gls{mp}($32$)-Pool-\gls{mp}($32$)-Unpool-\gls{mp}($32$)-Readout, with \gls{gin} as \gls{mp}. 
The Unpool operation (also referred to as \emph{lifting}~\citep{jin2020graph}) is implemented by copying in each node $i$ the value of the supernode $j$ to which the $i$ was assigned by the \texttt{SEL} operation in the pooling phase.
Zero-padding is used when lifting nodes not assigned to any supernode, like in the case of \gls{topk}.
Further details about the architecture for node classification and the unpooling procedure are deferred to Appendix~\ref{app:node_class}.

For this experiment, we considered the $5$ heterophilic datasets presented in~\cite{platonov2023a} (details in Appendix~\ref{app:node-classification-datasets}).
As pooling methods we considered \gls{topk}~\citep{gao2019graph}, \gls{kmis}~\citep{bacciu2023pooling}, \gls{ndp}~\citep{bianchi2020hierarchical}, and \gls{diffndp}.
We did not consider Graclus or any soft-clustering poolers, as they were exhausting the RAM and GPU VRAM, respectively, given the large size of the graphs. 
On the other hand, \gls{diffndp} is very parsimonious in terms of computational resources and scales very well with the graph size.
To systematically estimate the space complexity of the different pooling methods we performed an experimental evaluation of the GPU VRAM usage, which can be found in Appendix \ref{app:memory_usage}.
\input{tables_2025/node_class_res}

Following \cite{platonov2023a}, in Tab.~\ref{tab:node-classification} we report the means and standard deviations of the accuracy for Roman-empire and Amazon-ratings, and of the ROC AUC for Tolokers, Minesweeper, and Questions.
The results are computed on the $10$ public folds of these datasets.
When configured with \gls{diffndp} and \gls{diffndp}-E, the node classification architecture achieves significantly superior performance on the Roman-Empire dataset, which is notably the most heterophilic among all the datasets (see Tab.~\ref{tab:node-class_stats}). 
On the remaining datasets, our method performs well overall. 
Unlike the other pooling methods that achieve top performance only on a subset of the datasets, \gls{diffndp}-E is consistently in the top tier.

%%%%%%%%%%%%%%%%%%%
%%% CONCLUSIONS %%%
%%%%%%%%%%%%%%%%%%%
\section{Conclusion}

This work contributes significantly to both the \maxcut optimization and the development of specialized \gls{gnn} architectures to solve combinatorial optimization problems. 
Our proposed \gls{gnn}-based \maxcut algorithm not only extends the \maxcut optimization to attributed graphs and combines it with task-specific losses but also surpasses the performances of traditional methods on non-attributed graphs. 
While a conventional \gls{gnn} with a huge capacity manages to optimize a \maxcut loss~\citep{schuetz2022combinatorial}, our model is much more efficient thanks to the \acrlong{hetmp} layers.
This results highlight the importance of aligning the \gls{gnn} architecture with the problem's inherent structure: in this case, leveraging heterophilic propagation to solve problems that seek dissimilarity between neighboring nodes.

Our second contribution is to utilize the proposed \maxcut optimizer to implement a graph pooling method that combines the flexibility of soft-clustering approaches with the efficiency of scoring-based methods and with the theoretically-inspired design of one-every-$K$ strategies. \glspl{gnn} for graph and node classification equipped with our proposed pooling layer consistently achieves superior performance across diverse downstream tasks.
Differently from existing graph pooling and graph coarsening approaches that aim at preserving low-frequencies on the graph~\citep{loukas2019graph}, our method performs exceptionally well also on heterophilic datasets.
While our pooling layer can implement any pooling ratio, the auxiliary loss is optimized for the node partition induced by the \maxcut, whose size might not be aligned with the specified pooling ratio. When the distribution of the nodes' degree is approximately uniform, the \maxcut induces an approximately balanced partition corresponding to a pooling ratio of $\approx 0.5$, which is, thus, generally a good choice. 

Looking forward, we see great potential in pretraining \glspl{gnn} with auxiliary losses. This aligns with the principles of foundational models~\citep{bommasani2021opportunities} and could facilitate the development of more effective and general-purpose graph pooling techniques.

% If you'd like to, you may include a section for author contributions as is done in many journals. This is optional and at the discretion of the authors.

\subsubsection*{Acknowledgments}
This work was supported by the Norwegian Research Council project 345017: \emph{RELAY: Relational Deep Learning for Energy Analytics}. The authors wish to thank Nvidia Corporation for donating some of the GPUs used in this project.

\bibliography{biblio}
\bibliographystyle{iclr2025_conference}

\newpage
\appendix

\section*{\Large Appendix}

%%%%%%%%%%%%%%%%%
\section{Nearest neighbor association algorithm}
\label{app:nearest-neighbour}

As discussed in Sec.~\ref{sec:method}, each node is associated to one of the supernodes (preferably the closest in terms of path-distance on the graph).
Naively searching for each node the closest supernode is computationally demanding and becomes intractable for large graphs.
Therefore, we propose an implementation of the assignment scheme that is efficient and can be easily parallelized on a GPU.
The proposed algorithm is based on a Breadth First Search (BFS) of the graph and is detailed in the pseudo-code in Algorithm~\ref{alg:assignment}.

\input{algorithms/assignment}

The algorithm takes as input the graph $\gG$ (in particular, its topology described by the adjacency matrix $\mA$), the set of $K$ supernodes $\gS$ identified by the \texttt{SEL} operation, and a maximum number of iterations ($MaxIter$), which represent the maximum number of steps a node can traverse the graph to reach its closest supernode before being assigned at random.

In line~\ref{line:initE}, an encoding matrix $\mE$ of size $N \times K+1$ is initialized so that row $i$ is a one-hot vector with the non-zero entry in position $k+1$, if the node $i$ of the original graph is the $k$-th supernode. Otherwise, row $i$ in a zero-vector of size $K+1$. This matrix will be gradually populated when supernodes are encountered during the BFS. It's important to note that the $0$-th column in matrix $\mE$ (and subsequently in $\mE'$) serves a special purpose. This column represents a ``fake'' supernode, which plays a crucial role in the assignment process.

A Boolean mask $\vm \in \{0,1\}^N$ indicating whether a node already encountered the closest supernode is initialized in line~\ref{line:m} with $1$ in position $i$ is nodes $i$ is a supernode and $0$ otherwise.
Finally, an empty list indicating to which supernode each node is assigned is initialized (line~\ref{line:initA}).

Until the maximum number of iterations is reached or until all nodes are assigned (line~\ref{line:check}), the encoding matrix $\mE$ is propagated with an efficient message passing operation (line~\ref{line:mp}) that can be parallelized on a GPU. As soon as a $1$ appears in position $k$ within a line $i$ of $\mE$ previously full of zeros, node $i$ is assigned to supernode $k$ and the assignments and mask $\vm$ are updated accordingly (lines~\ref{line:updatea} and~\ref{line:updatem}). 
The \emph{ParallelAssignment} function (line~\ref{line:updatea}), in particular, takes the rows of the newly generated embeddings $\mE'$ that have not yet been assigned and performs an \texttt{argmax} operation on the last dimension. If the \texttt{argmax} doesn't find any valid supernode for a node (\emph{i.e.}, all values in the row are zero), it returns 0, effectively assigning the node to the ``fake'' supernode represented by the $0$-th column. This allows to filter out the unassigned nodes in line \ref{line:updatem}.

If there are still unassigned nodes at the end of the iterations, the remaining nodes are randomly assigned to one of the $K$ supernodes (line~\ref{line:randass}).
Finally, all the assignments are merged (line \ref{line:finala}).

%%%%%%%%%%%%%%%%%
\section{Derivation of the auxiliary loss}
\label{app:aux_loss}

Let us consider the \maxcut objective in Equation \ref{eq:maxcut}. 
It can be rewritten as 
\begin{align*}
    \max_{\vz} \Bigg( \sum \limits_{i,j \in  \mathcal{V}} w_{ij} - \sum \limits_{i,j \in  \mathcal{V}} z_i z_j w_{ij} \Bigg) = \max_{\vz} \Bigg( |E| - \sum \limits_{i,j \in  \mathcal{V}} z_i z_j w_{ij} \Bigg), \\
\end{align*}
which is equivalent to
\begin{equation*}
 \max_{\vz} \Bigg( 1 - \sum \limits_{i,j \in  \mathcal{V}} \frac{z_i z_j w_{ij}}{|E|} \Bigg).   
\end{equation*}
   
The solution $\vz^*$ for the original objective is thus the solution for 
$$
    \min_{\vz}
    % - \sum \limits_{i,j \in  \mathcal{V}} \frac{z_i z_j w_{ij}}{|E|} = 
    \frac{\vz^\top A \vz}{|E|}.
$$

%%%%%%%%%%%%%%%%%
\section{Implementation details}
\label{app:architectures}

%----------------
\subsection{MaxCutPool layer and ScoreNet}
\label{app:auxnet}

\begin{figure}[!ht]
    \centering
    \includegraphics[width=0.8\textwidth]{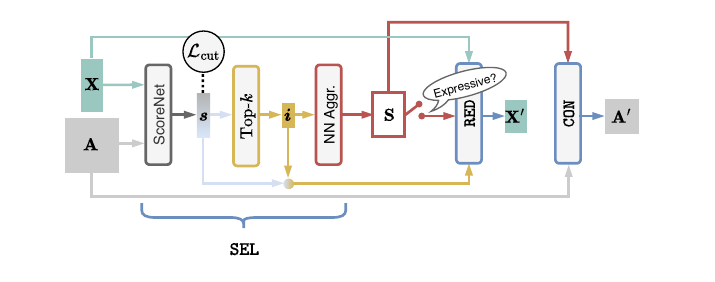}
    \caption{\small Scheme of the \gls{diffndp} layer.}
    \label{fig:maxcutpool_layer}
\end{figure}

A schematic depiction of the \gls{diffndp} layer is illustrated in Fig.~\ref{fig:maxcutpool_layer}, where the \texttt{SEL}, \texttt{RED}, and \texttt{CON} operations are highlighted.

\texttt{SEL} first computes a score vector $\boldsymbol{s}$ using the auxiliary GNN, ScoreNet, based on the node features $\mathbf{X}$ and the connectivity matrix $\mathbf{A}$.
A top-$K$ operation is used to find the indices $\boldsymbol{i}$ of the $K$ nodes with the highest scores that become the supernodes, \emph{i.e.}, the nodes of the pooled graph.
The remaining $N-K$ nodes are assigned to the nearest supernode through a nearest-neighbor (NN) aggregation procedure that yields an assignment matrix $\mathbf{S}$, whose $jk$-th element is $1$ if nodes $j$ is assigned to supernode $k$, and zero otherwise. 
The score vector $\boldsymbol{s}$ is used to compute the loss $\gL_\text{cut}$, which is associated with each \gls{diffndp} layer.

The \texttt{RED} operation computes the node features of the pooled graph $\mX'$ by multiplying the features of the selected nodes $\mX_i$ with the scores $\boldsymbol{s}$. 
This operation is necessary to let the gradients flow past the top-$K$ operation, which is not differentiable.
In the expressive variant, \gls{diffndp}-E, \texttt{RED} computes the new node features by combining those from all the nodes in the graph through the multiplication with matrix $\mS$.
We combine the features by summing them instead of taking the average since the sum enhances the expressiveness of the pooling layer~\citep{bianchi2023expr}.

The \texttt{CON} operations always leverage the assignment matrix to compute the adjacency matrix of the pooled graph.
In particular, the edge connecting two supernodes $i$ and $j$ is obtained by coalescing all the edges connecting the nodes assigned to supernode $i$ with those assigned to supernode $j$.
Also in this case, we take the sum as the operation to coalesce the edges.
The resulting edges in the pooled graph are associated with a weight $w_{ij}$ that counts the number of combined edges. 

\begin{figure}[!ht]
    \centering
    \includegraphics[width=0.7\textwidth]{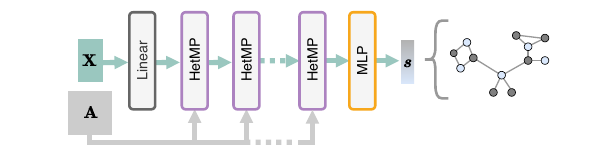}
    \caption{\small Scheme of the ScoreNet.}
    \label{fig:auxnet}
\end{figure}

The details of the ScoreNet used in the \gls{diffndp} layer are depicted in Fig.~\ref{fig:auxnet}.
The ScoreNet consists of a linear layer that maps the features $\mX$ to a desired hidden dimension.
Afterward, a stack of \gls{hetmp} layers gradually transforms the node features by amplifying their high-frequency components with heterogeneous \gls{mp} operations.
Finally, an \gls{mlp} transforms the node features of the last \gls{hetmp} layer into score vector $\vs$, which is a high-frequency graph signal.
We note that while the simple \gls{hetmp} we adopted works well in our case, different heterophilic \gls{mp} operators could have been considered~\citep{chien2021adaptive, dong2021adagnn, fu2022p}.

The ScoreNet is configured with the following hyperparameters:
\begin{itemize}
    \item Number of \gls{hetmp} layers and number of features in each layer. We use the notation $[32,16,8]$ to indicate a ScoreNet with three \gls{hetmp} layers with hidden sizes $32$, $16$, and $8$, respectively. We also use the notation $[32]\times 4$ to indicate $4$ layers with $32$ units each. As default, we use $[32, 32, 32, 32, 16, 16, 16, 16, 8, 8, 8, 8]$.
    \item Activation function of the \gls{hetmp} layers. As default, we use TanH.
    \item Number of layers and features in the \gls{mlp}. As default, we use $[16, 16]$.
    \item Activation function of the \gls{mlp}. As default, we use ReLU.
    \item Smoothness hyperparameter $\delta$. As default, we use $2$.
    \item Auxiliary loss weight $\beta$. As default, we use $1$.
\end{itemize}

The optimal configuration has been identified with the cross-validation procedure described in Sec.\ref{sec:experiments}. 
Depending on the experiment and the \gls{gnn} architecture, some parameters in the ScoreNet are kept fixed at their default value while others are optimized.

%----------------
\subsection{Cut model}
\label{app:cut_model}

\begin{figure}[!ht]
    \centering
    \includegraphics[width=0.6\textwidth]{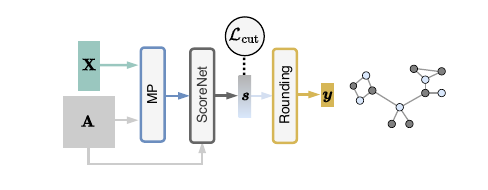}
    \caption{\small Scheme of the model used for computing the \maxcut.}
    \label{fig:cut_model}
\end{figure}

The model used to compute the \maxcut is depicted in Fig.~\ref{fig:cut_model}. The model consists of a single MP layer followed by the ScoreNet, which returns the score vector $\boldsymbol{s}$.
The \maxcut partition is obtained by rounding the values in the score vector as follows
\begin{equation*}
    y_i = 
    \begin{cases}
    1 & \text{if}\; s_i > 0, \\
    -1 & \text{otherwise}.
    \end{cases}
\end{equation*}
The model is trained in a completely unsupervised fashion only by minimizing the auxiliary loss $\gL_\text{cut}$.

As MP layers we used a GIN \citep{xu2018powerful} layer with $32$ units and ELU activation function. The model was trained for $2000$ epochs with Adam optimizer~\citep{DBLP:journals/corr/KingmaB14}, with the initial learning rate set to $8e-4$. We used a learning rate scheduler that reduces by $0.8$ the learning rate when the auxiliary loss does not improve for $100$ epochs. For testing, we restored the model checkpoint that achieved the lowest auxiliary loss.

The best configuration was found via a grid search on the following set of hyperparameters:
\begin{itemize}
    \item HetMP layers and units: 
    \begin{itemize}
        \item $[32] \times 4$, 
        \item $[4] \times 32$, 
        \item $[8] \times 16$, 
        \item $[16] \times 8$,
        \item $[32, 32, 32, 32, 16, 16, 16, 16, 8, 8, 8, 8]$.
    \end{itemize}
    \item HetMP activations: 
    \begin{itemize}
        \item ReLU, 
        \item TanH. 
    \end{itemize}
    \item Smoothness hyperparameter $\delta$:
    \begin{itemize}
        \item 2,
        \item 3,
        \item 5.
    \end{itemize}
\end{itemize}

In Tab. \ref{tab:cut_config} we report the configurations of the ScoreNet used for the different graphs in the \maxcut experiment.

\input{tables_2025/maxcut_hyperparams}

%----------------
\subsection{Graph classification model}
\label{app:graph_class}

\begin{figure}[!ht]
    \centering
    \includegraphics[width=0.6\textwidth]{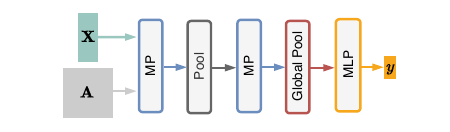}
    \caption{\small Scheme of the graph classification model.}
    \label{fig:graphclass_model}
\end{figure}

The model used to perform graph classification is depicted in Fig.~\ref{fig:graphclass_model}. 
The model consists of an \gls{mp} layer, followed by a pooling layer, an \gls{mp} acting on the pooled graph, a global pooling layer that sums the features of all the nodes in the pooled graph, and an \gls{mlp} that produces the label $y$ associated with the input graph.
The model is trained by jointly minimizing the cross-entropy loss between the predicted graph labels and the true ones and the auxiliary losses associated with the different pooling layers.
The models are trained using a batch size of $32$ for $1000$ epochs, using the Adam optimizer with an initial learning rate of $1e-4$. 
We used an early stopping that monitors the validation loss with a patience of $300$ epochs. For testing, we restored the model checkpoint that achieved the lowest validation loss during training.

The best configuration was found via a grid search on the following set of hyperparameters:
\begin{itemize}
    \item \gls{hetmp} layers and units: 
    \begin{itemize}
        \item $[32] \times 8$, 
        \item $[32] \times 4$, 
        \item $[8] \times 16$, $[16] \times 8$, 
        \item $[32,32,16,16,8,8]$,
        \item $[32, 32, 32, 32, 16, 16, 16, 16, 8, 8, 8, 8]$.
    \end{itemize}
    \item Auxiliary loss weight $\beta$: 
    \begin{itemize}
        \item 1,
        \item 2,
        \item 5.
    \end{itemize}
\end{itemize}

\input{tables_2025/graph_class_hp_merge}

In Tab. \ref{tab:graph_class_hp} we report the configurations of the ScoreNet used in the graph classification architecture for the different datasets in the expressive and non-expressive variant of \gls{diffndp}. 

%----------------
\subsection{Node classification model}
\label{app:node_class}

\begin{figure}[!ht]
    \centering
    \includegraphics[width=0.7\textwidth]{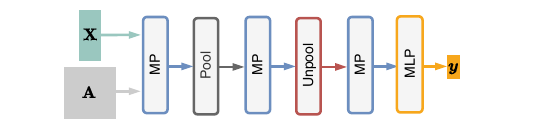}
    \caption{\small Scheme of the node classification model.}
    \label{fig:nodeclass_model}
\end{figure}

The model used to perform node classification is depicted in Fig.~\ref{fig:nodeclass_model}. 
The model consists of an \gls{mp} layer, followed by a pooling layer, an \gls{mp} acting on the pooled graph, an unpooling (lifting) layer, an \gls{mp} on the unpooled graph, and an \gls{mlp} that produces the final node labels $\boldsymbol{y}$.

The entry $y_i$ represents the predicted label for node $i$.
The model is trained by jointly minimizing the cross-entropy loss between the predicted node labels and the true ones and the auxiliary loss of the pooling layer.

\begin{figure}[!ht]
    \centering
    \includegraphics[width=0.8\textwidth]{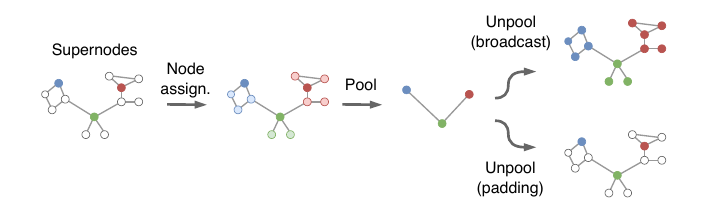}
    \caption{\small The two possible strategies for performing unpooling (lifting).}
    \label{fig:pool_padding}
\end{figure}

When using \gls{diffndp}, the unpooling/lifting procedure can be implemented in two different ways, illustrated in Fig.~\ref{fig:pool_padding}.
The first strategy, \emph{broadcast} unpooling, copies the values $\mX'$ of the nodes of the pooled graph to both the corresponding supernodes and to the nodes associated with the supernodes according to the assignment matrix $\mS$, obtained as described in Sec.~\ref{sec:method} and Appendix~\ref{app:nearest-neighbour}. Formally, the unpooled node features $\tilde{\mX}$ are: 
\begin{equation*}
    \tilde{\mX} = \mS \mX'.
\end{equation*}
We note that this is the commonly used approach to perform unpooling in cluster-based poolers.

In the second strategy, \emph{padding} unpooling, the values $\mX'$ are copied back only to the supernodes, while the remaining nodes are padded with a zero-valued vector:
\begin{equation*}
    [\tilde{\mX}]_i = 
    \begin{cases}
        [\mX']_i & \text{if $i$ is a supernode}\\
        \boldsymbol{0} & \text{otherwise}.
    \end{cases}
\end{equation*}
This is the approach to perform unpooling used by scoring-based approaches such as \gls{topk} and by one-over-$K$ approaches such as \gls{ndp} that only select the supernodes and leave the remaining nodes unassigned.

For the node classification task presented in Sec.~\ref{sec:node_class}, as \gls{mp} layers we used a GIN~\citep{xu2018powerful} layer with $32$ units and ReLU activation function.
The \gls{mlp} has a single hidden layer with $32$ units, a ReLU activation function, and a dropout layer between the hidden and output layers with a dropout probability of $0.1$.
The unpooling strategy used in this architecture is the broadcast one for \gls{kmis} and \gls{diffndp} and the padding one for \gls{topk} and \gls{ndp}.

The best configuration was found via a grid search on the following set of hyperparameters:
\begin{itemize}
    \item HetMP layers and units: 
    \begin{itemize}
        \item $[32] \times 4$, 
        \item $[4] \times 32$,
        \item $[32, 32, 32, 32, 16, 16, 16, 16, 8, 8, 8, 8]$.
    \end{itemize}
    \item MLP activations: 
    \begin{itemize}
        \item ReLU,
        \item TanH.
    \end{itemize}
\end{itemize}

The configuration of the ScoreNet for the \gls{diffndp} pooler used in the different datasets is reported in Tab.~\ref{tab:node-class_config}. 
\begin{table}[htbp]
\centering
\caption{Hyperparameters configurations of the ScoreNet in the node classification task.}
\label{tab:node-class_config}
\begin{tabular}{lcc}
\cmidrule[1.5pt]{1-3}
\textbf{Dataset}     & \gls{mp} units & \gls{mlp} Act. \\
\midrule
Roman-Empire         & $[32, 32, 32, 32]$ & ReLU \\
Amazon-Ratings       & $[32, 32, 32, 32]$ & ReLU \\
Minesweeper          & $[32, 32, 32, 32, 16, 16, 16, 16, 8, 8, 8, 8]$ & ReLU \\
Tolokers             & $[32, 32, 32, 32, 16, 16, 16, 16, 8, 8, 8, 8]$ & ReLU \\
Questions            & $[32, 32, 32, 32]$ & ReLU \\
\cmidrule[1.5pt]{1-3}
\end{tabular}
\end{table}

The node classifier was trained for $20,000$ epochs, using the Adam optimizer with an initial learning rate of $5e-4$. 
We used a learning rate scheduler that reduces by $0.5$ the learning rate when the validation loss does not improve for $500$ epochs.
We used an early stopping that monitors the validation loss with a patience of $2,000$ epochs. For testing, we restored the model checkpoint that achieved the lowest validation loss during training.

\begin{figure}[!ht]
    \centering
    \includegraphics[width=0.9\textwidth]{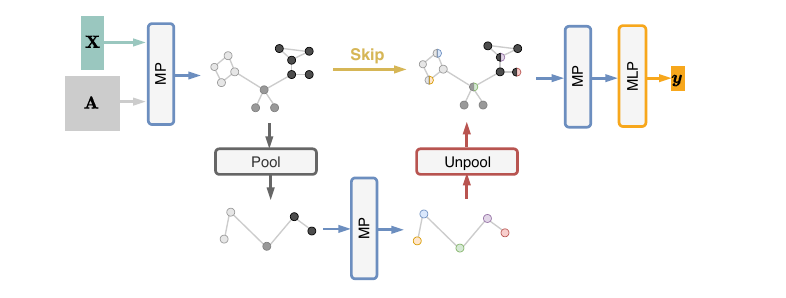}
    \caption{\small Scheme of the node classification model with skip connections.}
    \label{fig:nodeclasscat_model}
\end{figure}

We considered also an additional architecture for node classification with skip (residual) connections, depicted in Fig.~\ref{fig:nodeclasscat_model}.
This architecture is similar to the Graph U-Net proposed by~\cite{gao2019graph}.
The node features obtained after the first \gls{mp} layer are concatenated to the node features generated by the unpooling step.
In this architecture, we used the broadcast unpooling for \gls{kmis} and padding unpooling for \gls{diffndp}, \gls{topk}, and \gls{ndp}.
The results obtained with this architecture are reported in App.~\ref{app:additional-results}.
In Tab.~\ref{tab:node-class_config-cat} we report the configurations of the ScoreNet used in the architecture with skip connection in the different datasets. 
\begin{table}[htbp]
\centering
\caption{Hyperparameters configurations for the node classification task based on the architecture with skip connections.}
\label{tab:node-class_config-cat}
\begin{tabular}{lcc}
\cmidrule[1.5pt]{1-3}
\textbf{Dataset}     & \textbf{\gls{mp} units} & \textbf{\gls{mlp} Act.} \\
\midrule
Roman-Empire         & [32, 32, 32, 32] & ReLU \\
Amazon-Ratings       & [32, 32, 32, 32] & ReLU \\
Minesweeper          & [32, 32, 32, 32] & TanH \\
Tolokers             & [32, 32, 32, 32] & ReLU \\
Questions            & [32, 32, 32, 32] & ReLU \\
\cmidrule[1.5pt]{1-3}
\end{tabular}
\end{table}

%----------------
\subsection{Implementation of other pooling layers}\label{app:other_pooling}

The pooling methods \gls{topk}, Diffpool, \gls{dmon}, Graclus, and \acrlong{mincut} are taken from PyTorch Geometric~\citep{fey2019fast}.
For $k$-MIS we used the official implementation~\footnote{\url{https://github.com/flandolfi/k-mis-pool}}.
For \gls{ecpool}, we used the efficient parallel implementation~\footnote{\url{https://github.com/flandolfi/edge-pool}} proposed by~\cite{landolfi2022revisiting}.
For \gls{ndp} we adapted to PyTorch the original Tensorflow implementation~\footnote{\url{https://github.com/danielegrattarola/decimation-pooling}}.
All pooling layers were used with the default hyperparameters.
Since \gls{kmis} does not allow to directly specify the pooling ratio, we set $k = \floor{1/k}$.

%%%%%%%%%%%%%%%%%
\section{Datasets details}

%----------------
\subsection{Cut datasets}
\label{app:cut-datasets}

The statistics of the PyGSP and the Gset datasets used to compute the \maxcut partition in Sec.~\ref{sec:maxcut_experiment} are reported in Tab.~\ref{tab:maxcut_dataset}.
While the PyGSP graphs are built from the library~\citep{defferrard2017pygsp}, the Gset dataset is downloaded from the original source~\footnote{ \url{http://web.stanford.edu/~yyye/yyye/Gset/}}.

\input{tables_2025/maxcut_dataset}

%----------------
\subsection{Multipartite dataset description} 
\label{app:multipartite}

\input{algorithms/multipartite_v2}

The Multipartite graph dataset is a synthetic dataset consisting of complete multipartite graphs. The nodes of each graph can be partitioned into $C$ clusters of independent nodes, such that every node is connected to every node belonging to every other cluster. 
The generation of the graphs and the class labels is formally described by the pseudo-code in Algorithm~\ref{alg:multipartite} and is discussed in the following:

\begin{enumerate}
    \item A set of $C$ cluster centers with 2D coordinates $(x,y)$ is initially arranged in a polygon shape. Each center is associated with a label, \emph{i.e.}, a color.
    \item The graph class is determined by the position and the color of the cluster centers. Specifically, the graph class is given by the color of the cluster whose center is on the positive $x$-axis.
    \item For each class, we generate multiple graphs using these cluster centers. A graph is created by drawing at random the position of the nodes around each cluster center. The number of nodes per cluster varies randomly up to a maximum. Nodes within a cluster share the same color, which is determined by the cluster center.
    \item The topology of each graph is obtained by connecting nodes from one cluster to the nodes of all the other clusters, but not to the nodes of the same cluster. Therefore, a node is connected only to nodes with different colors, making the graphs highly heterophilic.
    \item After generating graphs for one class, the cluster centers are rotated, and this rotated configuration is used for the next class. Indeed, each rotation brings a different cluster to the positive $x$-axis.
    \item The rotation process continues until the graphs for all the $C$ different classes, whose number is equal to the number of clusters, are generated.
\end{enumerate}

\begin{figure}[hb]
    \centering
    \includegraphics[width=0.9\textwidth]{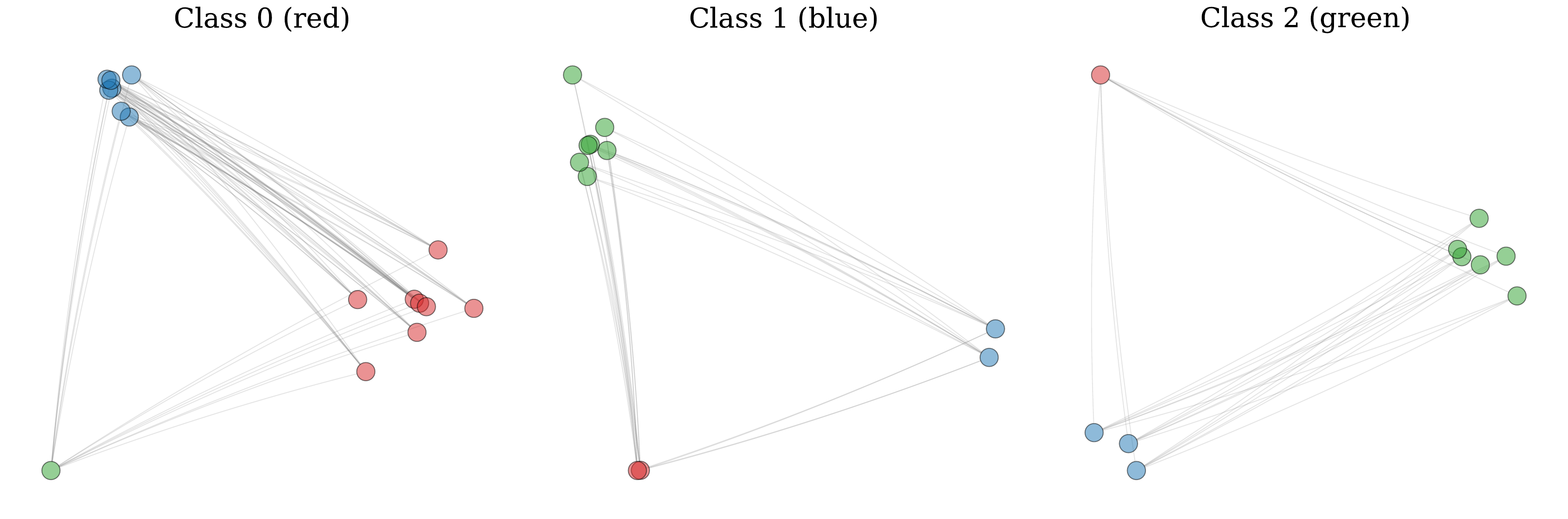}
    \caption{\small Example of multipartite graphs with $C=3$ cluster centers generated via our procedure. The graph class corresponds to the color of the nodes from the group to the right.}
    \label{fig:multipartite}
\end{figure}

Examples of multipartite graphs obtained for $C=3$ are shown in Figure \ref{fig:multipartite}.

The process depends on a few parameters that determine the number of clusters, the maximum nodes per cluster, and the number of graphs per class, providing control over the dataset size and complexity. The specific instance of the dataset used in our experiments has $10$ centers, $500$ graphs per center, and a maximum of $20$ nodes per cluster, and is available online~\footnote{
\url{https://zenodo.org/doi/10.5281/zenodo.11616515}
%\url{hidden_for_peer_review}
}.

The Multipartite dataset is intentionally designed so that the class label is determined solely by node features: specifically, the color and one of the 2D coordinates (the node's position along the $x$-axis). Although the graph's topology is structured to ensure that each graph is multipartite, this structure is independent of the class label. This creates an intriguing dichotomy between the graph's topology and its classification labels. 
In theory, a simple \gls{mlp} focusing exclusively on node features could accurately solve the classification task, as the graph's topology is essentially irrelevant for determining the correct labels. However, when processed by \glspl{gnn}, this dataset allows us to explore whether the model can correctly identify and utilize the relevant node features for classification, despite the presence of potentially misleading or noisy topological information. 
Through this carefully constructed dataset, we aim to highlight the strengths and potential limitations of certain \glspl{gnn} architectures, particularly in scenarios where the relationship between graph structure and classification labels is non-trivial, such as in heterophilic datasets.

%----------------
\subsection{Graph classification datasets} 
\label{app:graph-classification-datasets}

In addition to the novel Multipartite dataset introduced in Sec.~\ref{app:multipartite}, we consider $10$ datasets for graph classification in our experimental evaluation.
The TU Datasets~\citep{morris2020tudataset} (NCI, PROTEINS, Mutagenicity, COLLAB, REDDIT-B, DD, MUTAG, and Enzymes) are obtained through the loader of PyTorch Geometric~\footnote{\url{https://pytorch-geometric.readthedocs.io/en/latest/generated/torch_geometric.datasets.TUDataset.html}}.
The EXPWL1 and GCB-H datasets, respectively introduced by \cite{bianchi2023expr} and by~\cite{bianchi2022pyramidal}, are taken from the official repositories~\footnote{\url{https://github.com/FilippoMB/The-expressive-power-of-pooling-in-GNNs}}~\footnote{\url{https://github.com/FilippoMB/Benchmark_dataset_for_graph_classification}}.
The statistics of each dataset are reported in Tab.~\ref{tab:gc_dataset}.

\input{tables_2025/graph_class_stats}

Since the node labels are not available in the graph classification setting, it is not possible to rely on the homophily ratio $h(\gG)$~\citep{lim2021large} considered in the node classification setting.
Therefore, to quantify the degree of homophily in the graphs we look at the node features instead and introduce a surrogate homophily score $\bar h (\gD)$, where $\gD$ denotes the whole dataset.
The new score is defined as the absolute value of the average cosine similarity between the node features of connected nodes in each graph of the dataset:
\begin{equation*}
   \bar h(\gD) = \Bigg\lvert \frac{1}{|\gD|} \sum_{\gG \in \gD} \frac{1}{|\gE_\gG|} \sum_{(i,j) \in \gE_\gG}  \frac{\vx_i \vx_j}{ \| \vx_i\| \|\vx_j\|} \Bigg\rvert
\end{equation*}
where $|\gD|$ is the number of graphs in the dataset, $\gE_\gG$ is the set of edges of the graph $\gG$ and $\vx_i, \vx_j$ are the feature vectors of the $i$-th and $j$-th node respectively.

%----------------
\subsection{Node classification datasets}
\label{app:node-classification-datasets}

The datasets are the heterophilic graphs introduced by~\cite{platonov2023a} and are loaded with the API provided by PyTorch Geometric~\footnote{\url{https://pytorch-geometric.readthedocs.io/en/latest/generated/torch_geometric.datasets.HeterophilousGraphDataset.html}}.
The nodes of each graph are already split in train, validation, and test across 10 different folds.
The statistics of the five datasets are reported in Tab.~\ref{tab:node-class_stats}.
\input{tables_2025/node_class_stats}
The column $h(\gG)$ is the class insensitive edge homophily ratio as defied by~\cite{lim2021large}, which represents a measure for the level of homophily in the graph.

%%%%%%%%%%%%%%%%%
\section{Additional results}
\label{app:additional-results}

\subsection{Graph classification}
\label{app:graph_class_extra}
In Tab.~\ref{tab:graph-classification-subset} we report the additional graph classification results for the dataset where GNNs equipped with different pooling operators did not achieve a significantly different performance from each other.

\input{tables_2025/graph_class_res_remaining}

\subsection{Node classification}
\label{app:node_class_extra}
Tab.~\ref{tab:node-classification-cat} presents the results for node classification using the architecture with skip connections described in Appendix~\ref{app:node_class}. For this architecture, we focused on the non-expressive variant of \gls{diffndp}, which consistently delivered superior performance in this context.
The improved results can be attributed to the architecture's ability to preserve the original node information through skip connections. Additionally, by avoiding the combination of neighboring node features (as is done in the expressive variant) the model is better equipped to learn high-frequency features, which is particularly advantageous for heterophilic datasets.
For the Minesweeper dataset, we chose to use a GIN layer with $16$ units as the \gls{mp} layer, instead of the usual $32$ units. This decision was made because, regardless of the pooling method used, the architecture with skip connections consistently achieved nearly $100$\% ROC AUC whenever configured with a higher capacity.
\input{tables_2025/node_class_cat_res}
\section{Complexity}

We first discuss the algorithmic complexities and then report empirical measurements about processing time and memory usage.
All measurements are done on an Nvidia RTX A6000.

\subsection{Algorithmic complexity}
\label{app:algo_comp}

The complexity of \gls{diffndp} depends on the complexities of the operations \texttt{SEL}, \texttt{RED}, and \texttt{CON} and of the auxiliary loss $\mathcal{L}_\text{cut}$.

\paragraph{\texttt{SEL}}
The complexity of the \texttt{SEL} operation depends on the ScoreNet, which consists of a stack of $L$ \gls{hetmp} layers followed by an \gls{mlp}, and on the $\text{top}_K$ selection.

\begin{itemize}
    \item \textbf{\gls{hetmp}.}
    Following the analysis in \cite{blakely2019complexity}, for a graph with $N$ nodes, $E$ edges, and $F$ features, each \gls{hetmp} layer has a time complexity of $\mathcal{O}(NF^2 + EF)$ and a space complexity of $\mathcal{O}(E + NF + F^2)$. This results in a space and time complexity of $\mathcal{O}(N + E)$ with respect to the input.
    \item \textbf{\gls{mlp}.}
    The \gls{mlp} has a fixed structure with predetermined layer sizes. Since it operates independently on each node's feature vector and the number of operations per node is constant, processing each node takes $\mathcal{O}(1)$ time. With $N$ nodes to process, this results in a total time complexity of $\mathcal{O}(N)$ with respect to the input. The space complexity is also $\mathcal{O}(N)$, as we need to store the \gls{mlp} hidden and output features for each node.
    \item \textbf{top}$_K$
    The complexity of sorting an array of $N$ elements is $\mathcal{O}(N\log(N))$. However, if we are interested in finding only the top-$K$ elements the complexity can be lowered to $\mathcal{O}(N\log(K))$ or $\mathcal{O}(N + K)$, depending on the algorithm adopted. Therefore, we can assume an almost-linear complexity in time and space with respect to the number of nodes $N$.
\end{itemize}

\paragraph{\texttt{RED}}
For the non-expressive variant, \texttt{RED} involves a Hadamard product between the scores and features of the $K$ selected nodes, giving a time complexity of $\mathcal{O}(K)$. The expressive variant requires an additional multiplication with the assignment matrix $\mS$, increasing the complexity to $\mathcal{O}(K + NF)$. When $K$ is a function of $N$ (e.g., $K=N/2$), both variants have a time complexity of $\mathcal{O}(N)$ with respect to the input.

The space complexity is $\mathcal{O}(N)$, representing the storage of input and output data.

\paragraph{\texttt{CON}}
Our efficient implementation of the nearest neighbor assignment follows the complexity of BFS: $\mathcal{O}(N + E)$ time and $\mathcal{O}(N)$ space.

\paragraph{Auxiliary loss}
The auxiliary loss $\mathcal{L}_\text{cut}$ requires computing a quadratic form, with time complexity $\mathcal{O}(E)$ and space complexity $\mathcal{O}(N+E)$.

\paragraph{Total complexity}
The overall complexity of \gls{diffndp} is:

\begin{equation}
    \label{app:space-time-complex}
    \begin{aligned}
    \text{Time complexity:} &  \;\;\mathcal{O}(E + N) \\
    \text{Space complexity:} & \;\; \mathcal{O}(E + N)
    \end{aligned}
\end{equation}

These sub-quadratic complexities match those of the most efficient \gls{mp} and trainable pooling operators.

\subsection{Execution times}
\label{app:time_comp}

In Tab.~\ref{tab:exec_times} we report the number of seconds used by the architecture for node classification to process a batch when configured with different pooling operators.

\input{tables_2025/execution_times}

We note that one-over-$K$ methods such as Graclus, \gls{ndp}, and \gls{kmis} perform a preprocessing step on the CPU before the training starts.
Such operations are not accounted for in the measurements in Tab.~\ref{tab:exec_times}, but they can take significant time and be a bottleneck in those cases that require operations such as eigenvalues decomposition.

\subsection{Memory usage}
\label{app:memory_usage}

\input{tables_2025/memory_usage}

\begin{figure}[!ht]
    \centering
    \includegraphics[width=0.8\textwidth]{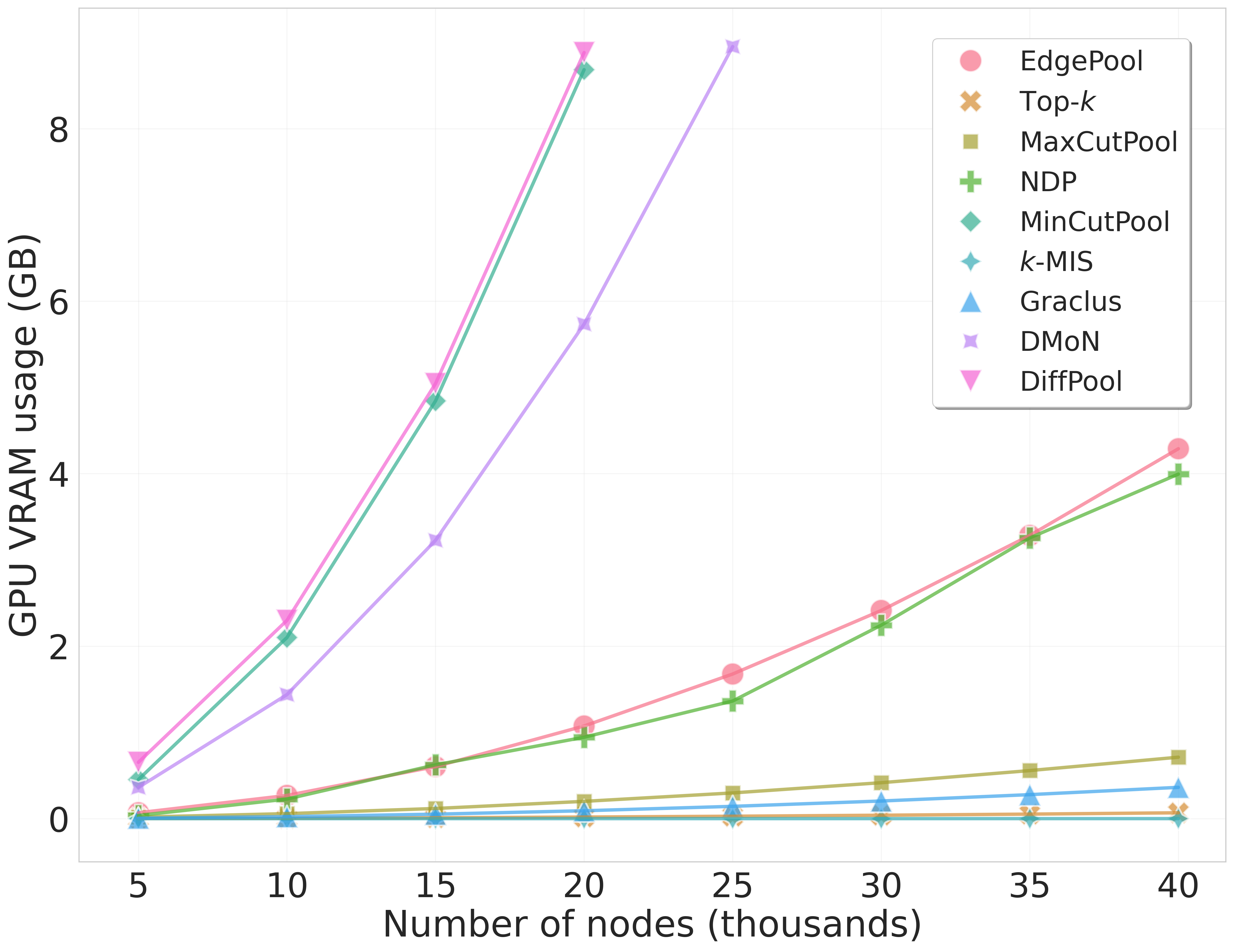}
    \caption{\small The GPU VRAM usage of the different poolers.}
    \label{fig:gpu_usage_plot}
\end{figure}

In Tab.~\ref{tab:mem_usage} we report the average and maximum GPU VRAM used by the architecture for node classification on the different datasets. As for the time, the values reported for Graclus, \gls{ndp}, and \gls{kmis} do not include the \texttt{SEL} and \texttt{CON} operations performed in preprocessing. 

To give a more interpretable demonstration of how the space complexity scales for the different pooling methods, in Fig.~\ref{fig:gpu_usage_plot} we report the GPU VRAM usage of the different poolers when processing a randomly generated Erdős-Renyi graph of increasing size.
All the graphs are generated keeping at $0.01$ the probability of having an edge between any pair of nodes.

The plot shows that in soft-clustering methods the GPU VRAM usage grows exponentially with the graph size.
On the other hand, for scoring-based methods, including the proposed \gls{diffndp}, the growth is sublinear, making these approaches extremely suitable for working with large graphs.

\end{document}

%% file: math_commands.tex
\usepackage{amsmath,amsfonts,bm}

\def\eqref#1{equation~\ref{#1}}
\def\floor#1{\lfloor #1 \rfloor}
\def\1{\bm{1}}

\def\vi{{\bm{i}}}

\def\vm{{\bm{m}}}

\def\vs{{\bm{s}}}

\def\vu{{\bm{u}}}

\def\vx{{\bm{x}}}

\def\vz{{\bm{z}}}

\def\mA{{\bm{A}}}

\def\mD{{\bm{D}}}
\def\mE{{\bm{E}}}

\def\mI{{\bm{I}}}

\def\mL{{\bm{L}}}

\def\mP{{\bm{P}}}

\def\mS{{\bm{S}}}

\def\mX{{\bm{X}}}

\DeclareMathAlphabet{\mathsfit}{\encodingdefault}{\sfdefault}{m}{sl}
\SetMathAlphabet{\mathsfit}{bold}{\encodingdefault}{\sfdefault}{bx}{n}

\def\gD{{\mathcal{D}}}
\def\gE{{\mathcal{E}}}

\def\gG{{\mathcal{G}}}

\def\gL{{\mathcal{L}}}

\def\gS{{\mathcal{S}}}

\def\gV{{\mathcal{V}}}

\def\sR{{\mathbb{R}}}

\newcommand{\R}{\mathbb{R}}

%% file: notation.tex
\usepackage{amsmath, amsfonts, bm}
\usepackage{xspace}
\usepackage{pifont}
\newcommand{\xmark}{\ding{55}}

\def\eqref#1{equation~\ref{#1}}
\def\floor#1{\lfloor #1 \rfloor}
\def\1{\bm{1}}
\def\vi{{\bm{i}}}

\def\vm{{\bm{m}}}

\def\vs{{\bm{s}}}

\def\vu{{\bm{u}}}

\def\vx{{\bm{x}}}

\def\vz{{\bm{z}}}

\def\mA{{\bm{A}}}

\def\mD{{\bm{D}}}
\def\mE{{\bm{E}}}

\def\mI{{\bm{I}}}

\def\mL{{\bm{L}}}

\def\mP{{\bm{P}}}

\def\mS{{\bm{S}}}

\def\mX{{\bm{X}}}

\DeclareMathAlphabet{\mathsfit}{\encodingdefault}{\sfdefault}{m}{sl}
\SetMathAlphabet{\mathsfit}{bold}{\encodingdefault}{\sfdefault}{bx}{n}
\def\gD{{\mathcal{D}}}
\def\gE{{\mathcal{E}}}

\def\gG{{\mathcal{G}}}

\def\gL{{\mathcal{L}}}

\def\gS{{\mathcal{S}}}

\def\gV{{\mathcal{V}}}

\def\sR{{\mathbb{R}}}

\newcommand{\geqzero}{\scalebox{.7}{$\geq 0$}}
\DeclareMathSymbol{\shortminus}{\mathbin}{AMSa}{"39}

\newcommand{\maxcut}{\texttt{\small{MAXCUT}}\xspace}
\newcommand{\mincut}{\texttt{\small{minCUT}}\xspace}

%% file: acronyms.tex
\usepackage[acronym, nohypertypes={acronym}]{glossaries}
\glsdisablehyper

\newacronym{gsp}{GSP}{Graph Signal Processing}
\newacronym{src}{SRC}{Select-Reduce-Connect}
\newacronym{gw}{GW}{Goeman-Williamson}
\newacronym{levs}{LEVS}{largest eigenvector vertex selection}

\newacronym{cnn}{CNN}{Convolutional Neural Network}
\newacronym{gnn}{GNN}{Graph Neural Network}
\newacronym{mlp}{MLP}{Multilayer Perceptron}
\newacronym{gcn}{GCN}{Graph Convolutional Network}
\newacronym{gin}{GIN}{Graph Isomorphism Network}

\newacronym{mp}{MP}{Message Passing}
\newacronym{ndp}{NDP}{Node Decimation Pooling}
\newacronym{mincut}{MinCutPool}{MinCutPool}
\newacronym{topk}{Top-$k$}{Top-$k$ Pooling}
\newacronym{dmon}{DMoN}{Deep Modularity Networks}
\newacronym{kmis}{$k$-MIS}{$k$ Maximal Independent Sets Pooling}
\newacronym{ogcn}{$\omega$-GCN}{$\omega$-Graph Convolution Network}
\newacronym{diffndp}{MaxCutPool}{}
\newacronym{ecpool}{ECPool}{Edge-Contraction Pooling}
\newacronym{hetmp}{HetMP}{Heterophilic Message Passing}

\newacronym{jbgnn}{JBGNN}{Just-balance Graph Neural Network}

\newacronym{ogb}{ogbg-ppa}{protein-protein association networks}
\newacronym{gcb}{GCB-H}{Graph Classification Benchmark Hard}

%% file: tables_2025/pooling_drawbacks.tex
\bgroup
\setlength\tabcolsep{.95em} %horizontal padding
\begin{table}[htbp]
\centering
\caption{Drawbacks of different types of pooling operators.}
\label{tab:drawbacks}
\begin{tabular}{@{}lll@{}}
\textbf{Soft-clustering} & \textbf{Score-based} & \textbf{One-every-$\boldsymbol{K}$} \\
\toprule
\xmark \; Not adaptive to graph size & \xmark \; Pooling not uniform & \xmark \; Limited flexibility \\
\xmark \; Dense and not interpretable pooled graphs & \xmark \; Not expressive & \xmark \; Features agnostic \\
\xmark \; High memory cost & \xmark \; Worse performance &  \xmark \; Task agnostic \\
\bottomrule
\end{tabular}
\end{table}
\egroup

%% file: tables_2025/maxcut_res.tex
\bgroup
\def\arraystretch{1.0} %vertical padding 
\setlength\tabcolsep{.2em} %horizontal padding
\begin{table}[ht]
\fontfamily{bch}\selectfont
\caption{Size of the graph cuts obtained with \gls{diffndp}, a \gls{gnn} with \gls{gcn} layers, and two common algorithms to compute the \maxcut. GW results are absent for some entries of the PyGSP datasets and for GSet because the solver failed to converge.}
\label{tab:maxcut}
\centering
\subfloat[PyGSP datasets]{

\begin{tabular}{@{}l|cccc@{}}
\cmidrule[1.5pt]{1-5}
Dataset        & GW              & NDP             & GCN    & \gls{diffndp} \\ \midrule
BarabasiAlbert & 0.6875          & 0.6589          & 0.7240 & \textbf{0.7292}                \\
Community      & 0.6767          & 0.6429          & 0.6805 & \textbf{0.6814}                \\
ErdősRenyi     & 0.6920          & 0.6858          & 0.6797 & \textbf{0.7105}                \\
Grid (10$\times$10) & \textbf{1.0000} & \textbf{1.0000} & 0.9222 & \textbf{1.0000}                \\
Grid (60$\times$40) & -               & 0.9787          & 0.1862 & \textbf{0.9815}                \\
Minnesota      & -               & 0.9104          & 0.8904 & \textbf{0.9130}                \\
RandRegular  & 0.4827               & 0.8760          & 0.8733 & \textbf{0.9040}                \\
Ring           & \textbf{1.0000}               & \textbf{1.0000} & 0.4200 & \textbf{1.0000}                \\
Sensor         & 0.6000          & 0.5719          & 0.6281 & \textbf{0.6406}                \\ 
\cmidrule[1.5pt]{1-5}
\end{tabular}
  }
\subfloat[GSet datasets]{
\begin{tabular}{@{}l|ccc@{}}
\cmidrule[1.5pt]{1-4}
Dataset & NDP             & GCN    & \gls{diffndp} \\ \midrule
G14     & 0.6155          & 0.6323 & \textbf{0.6412}                \\
G15     & 0.5945          & 0.6288 & \textbf{0.6424}                \\
G22     & 0.6441          & 0.6409 & \textbf{0.6577}                \\
G49     & \textbf{1.0000} & 0.9683 & \textbf{1.0000}                \\
G50     & \textbf{0.9800} & 0.9610 & 0.9750                         \\
G55     & 0.7568          & 0.7865 & \textbf{0.8068}                \\
G70     & 0.8803          & 0.8945 & \textbf{0.9086}                \\ \cmidrule[1.5pt]{1-4}
\end{tabular}
}
\end{table}
\egroup

%% file: tables_2025/graph_class_student_fixed.tex
\bgroup
\def\arraystretch{.8} %vertical padding
\setlength\tabcolsep{.3em} %horizontal padding
\begin{table}[ht]
\caption{Mean and standard deviations of the graph classification accuracy. For each dataset the best performing method and those that are not significantly different from it are colored in green.
If a method is in the top-performing group is assigned with a score of $1$, $0$ otherwise.}
\label{tab:graph-classification}
\makebox[\textwidth]{
\fontfamily{bch}\selectfont
\centering
\begin{tabular}{@{}l|ccccccc|c@{}}
\cmidrule[1.5pt]{1-9}
\textbf{Pooler}  & \textbf{GCB-H} & \textbf{COLLAB} & \textbf{EXPWL1} & \textbf{Mult.} & \textbf{Mutag.} & \textbf{NCI1} & \textbf{REDDIT-B} & \textbf{Score} \\
\midrule[1pt]
No pool     & \textcolor{gray}{\textit{74{{\tiny$\pm$4}}}} & \textcolor{gray}{\textit{74{{\tiny$\pm$2}}}} & \textcolor{gray}{\textit{ 87{{\tiny$\pm$2}}}}  & \textcolor{gray}{\textit{ 14{{\tiny$\pm$12}}}} & \textcolor{gray}{\textit{79{{\tiny$\pm$2}}}} & \textcolor{gray}{\textit{78{{\tiny$\pm$3}}}} & \textcolor{gray}{\textit{90{{\tiny$\pm$2}}}} & -\\ 
\midrule
DiffPool     & \textcolor{black}{51{{\tiny$\pm$8}}} & \textcolor{black}{70{{\tiny$\pm$2}}} & \textcolor{black}{69{{\tiny$\pm$3}}}  & \textcolor{black}{9{{\tiny$\pm$1}}}   & 78{{\tiny$\pm$2}} & 75{{\tiny$\pm$2}} & \textcolor{darkgreen}{90{{\tiny$\pm$2}}} & 1 \\
\gls{dmon}         & \textcolor{darkgreen}{74{{\tiny$\pm$3}}} & \textcolor{black}{68{{\tiny$\pm$2}}} & \textcolor{black}{73{{\tiny$\pm$3}}}  & \textcolor{black}{52{{\tiny$\pm$2}}}  & \textcolor{darkgreen}{80{{\tiny$\pm$2}}} & \textcolor{darkgreen}{77{{\tiny$\pm$2}}} & 88{{\tiny$\pm$2}}& 3 \\
EdgePool     & \textcolor{darkgreen}{75{{\tiny$\pm$4}}} & \textcolor{black}{72{{\tiny$\pm$3}}} & 90{{\tiny$\pm$2}}  & 55{{\tiny$\pm$3}}  & \textcolor{darkgreen}{80{{\tiny$\pm$2}}} & \textcolor{darkgreen}{77{{\tiny$\pm$3}}} & \textcolor{darkgreen}{91{{\tiny$\pm$2}}} & 4 \\
Graclus      & \textcolor{darkgreen}{75{{\tiny$\pm$3}}} & \textcolor{black}{72{{\tiny$\pm$3}}} & 90{{\tiny$\pm$2}}  & \textcolor{black}{25{{\tiny$\pm$18}}} & \textcolor{darkgreen}{80{{\tiny$\pm$2}}} & \textcolor{darkgreen}{77{{\tiny$\pm$2}}} & \textcolor{darkgreen}{90{{\tiny$\pm$3}}} & 4 \\
\gls{kmis}        & \textcolor{darkgreen}{75{{\tiny$\pm$4}}} & \textcolor{black}{71{{\tiny$\pm$2}}} & \textcolor{darkgreen}{99{{\tiny$\pm$1}}}  & 58{{\tiny$\pm$2}} & \textcolor{darkgreen}{79{{\tiny$\pm$2}}} & 75{{\tiny$\pm$3}} & \textcolor{darkgreen}{90{{\tiny$\pm$2}}} & 4 \\
\acrlong{mincut}       & \textcolor{darkgreen}{75{{\tiny$\pm$5}}} & \textcolor{black}{70{{\tiny$\pm$2}}} & \textcolor{black}{71{{\tiny$\pm$3}}}  & 56{{\tiny$\pm$3}}  & 78{{\tiny$\pm$3}} & \textcolor{black}{73{{\tiny$\pm$3}}} & \textcolor{black}{87{{\tiny$\pm$2}}} & 1 \\
\gls{topk}         & \textcolor{black}{56{{\tiny$\pm$5}}} & \textcolor{black}{72{{\tiny$\pm$2}}} & \textcolor{black}{73{{\tiny$\pm$2}}}  & \textcolor{black}{43{{\tiny$\pm$3}}}  & \textcolor{black}{75{{\tiny$\pm$3}}} & \textcolor{black}{73{{\tiny$\pm$2}}} & \textcolor{black}{77{{\tiny$\pm$2}}} & 0\\
\midrule
\gls{diffndp}      & \textcolor{darkgreen}{73{{\tiny$\pm$3}}} & \textcolor{darkgreen}{77{{\tiny$\pm$2}}} & \textcolor{darkgreen}{100{{\tiny$\pm$0}}} & \textcolor{darkgreen}{90{{\tiny$\pm$2}}}  & 77{{\tiny$\pm$2}} & 75{{\tiny$\pm$2}} & \textcolor{darkgreen}{89{{\tiny$\pm$3}}} & 5 \\
\gls{diffndp}-E & \textcolor{darkgreen}{74{{\tiny$\pm$3}}} & \textcolor{darkgreen}{77{{\tiny$\pm$2}}} & \textcolor{darkgreen}{100{{\tiny$\pm$0}}} & \textcolor{darkgreen}{87{{\tiny$\pm$5}}}  & \textcolor{darkgreen}{79{{\tiny$\pm$1}}} & \textcolor{darkgreen}{76{{\tiny$\pm$2}}} & \textcolor{darkgreen}{89{{\tiny$\pm$2}}} & \textbf{7}\\ 
\midrule
\gls{diffndp}-NL & \textcolor{black}{61{{\tiny$\pm$6}}} & \textcolor{darkgreen}{77{{\tiny$\pm$3}}} & \textcolor{darkgreen}{100{{\tiny$\pm$0}}} & \textcolor{darkgreen}{91{{\tiny$\pm$1}}} & \textcolor{black}{76{{\tiny$\pm$3}}} & \textcolor{black}{74{{\tiny$\pm$2}}} & \textcolor{black}{86{{\tiny$\pm$3}}} & 3 \\
\cmidrule[1.5pt]{1-9}
\end{tabular}
}

\end{table}
\egroup

%% file: tables_2025/node_class_res.tex
\bgroup
\def\arraystretch{.8} %vertical padding
\setlength\tabcolsep{.5em} %horizontal padding
\begin{table}[ht]
\caption{Node classification accuracy (Roman-empire, Amazon-ratings) and AUROC (Minesweeper, Tolokers, Questions). The best performing models in each dataset are in green and get 1 score point, 0 otherwise.}
\label{tab:node-classification}
\makebox[\textwidth]{
\fontfamily{bch}\selectfont
\centering
\begin{tabular}{@{}l|ccccc|c@{}}
\cmidrule[1.5pt]{1-7}
\textbf{Pooler} & \textbf{Roman-e.} & \textbf{Amazon-r.} & \textbf{Minesw.} & \textbf{Tolokers} & \textbf{Questions} & \textbf{Score} \\ 
\midrule
No pool 
& \textcolor{gray}{\textit{59{{\tiny$\pm$0}}}} 
& \textcolor{gray}{\textit{46{{\tiny$\pm$1}}}} 
& \textcolor{gray}{\textit{86{{\tiny$\pm$2}}}}  
& \textcolor{gray}{\textit{86{{\tiny$\pm$4}}}} 
& \textcolor{gray}{\textit{71{{\tiny$\pm$2}}}} & - \\
% No pool 2
% & \textcolor{gray}{\textit{74{{\tiny$\pm$1}}}} 
% & \textcolor{gray}{\textit{53{{\tiny$\pm$1}}}} 
% & \textcolor{gray}{\textit{98{{\tiny$\pm$0}}}}  
% & \textcolor{gray}{\textit{89{{\tiny$\pm$0}}}} 
% & \textcolor{gray}{\textit{70{{\tiny$\pm$2}}}} & - \\
\midrule
\gls{topk}      & 26{\tiny$\pm$7} & 46{\tiny$\pm$4} & 94{\tiny$\pm$1} & \textcolor{darkgreen}{89{\tiny$\pm$5}} & 64{\tiny$\pm$3} & 1 \\
\gls{kmis}      & 23{\tiny$\pm$3} & 48{\tiny$\pm$2} & 75{\tiny$\pm$2} & 84{\tiny$\pm$2} & \textcolor{darkgreen}{83{\tiny$\pm$1}} & 1 \\
\gls{ndp}       & 22{\tiny$\pm$5} & \textcolor{darkgreen}{53{\tiny$\pm$2}} & \textcolor{darkgreen}{98{\tiny$\pm$0}} & \textcolor{darkgreen}{88{\tiny$\pm$6}} & 68{\tiny$\pm$4} & 3 \\
\gls{diffndp}   & \textcolor{darkgreen}{56{\tiny$\pm$3}} & \textcolor{darkgreen}{53{\tiny$\pm$1}} & \textcolor{darkgreen}{96{\tiny$\pm$1}} & 87{\tiny$\pm$3} & \textcolor{darkgreen}{82{\tiny$\pm$4}} & 4 \\
\gls{diffndp}-E & \textcolor{darkgreen}{60{\tiny$\pm$4}} & \textcolor{darkgreen}{53{\tiny$\pm$2}} & \textcolor{darkgreen}{97{\tiny$\pm$1}} & \textcolor{darkgreen}{91{\tiny$\pm$2}} & \textcolor{darkgreen}{85{\tiny$\pm$5}} & \textbf{5} \\
\cmidrule[1.5pt]{1-7}
\end{tabular}
}

\end{table}
\egroup

%% file: algorithms/assignment.tex
\begin{algorithm}
\caption{Pseudo-code for the assignment scheme to the supernodes}
\begin{algorithmic}[1]
\Procedure{AssignNodesToSupernodes}{$\gG, \gS, MaxIter$}
    \State $\mE \gets \text{InitializeEncodings}(\gG, \gS)$ \Comment{One-hot encoding} \label{line:initE}
    \State $\vm \gets \text{InitializeMask}(\gG, \gS)$ \label{line:m}
    \State $Assignments \gets \text{InitializeEmptyList}()$ \label{line:initA}
    \For{$i = 1$ to $MaxIter$}
        \If{$\text{AllNodesAssigned}(\vm)$} \label{line:check}
            \State \textbf{break}
        \EndIf
        \State $\mE' \gets \text{ParallelMessagePassing}(\gG, \mE)$
        \Comment{$\mE' = \mA\mE$} \label{line:mp}
        \State $Assignments \gets \text{ParallelAssignment}(\mE', \gS, \vm)$ \label{line:updatea}
        \State $\vm \gets \text{UpdateMask}(\vm, Assignments)$ \label{line:updatem}
        \State $\mE \gets \mE'$
    \EndFor
    \If{$\text{not AllNodesAssigned}(\vm)$}
        \State $RndAssignments \gets \text{ParallelRandomAssignment}(UnassignedNodes, \gS)$ \label{line:randass}
        %\State $\vm \gets \text{UpdateMask}(\vm, RndAssignments)$
    \EndIf
    \State $FinalAssignments \gets \text{GetFinalAssignments}(Assignments, RndAssignments)$
    \label{line:finala}
    \State \textbf{return} $FinalAssignments$
\EndProcedure
\end{algorithmic}
\label{alg:assignment}
\end{algorithm}

%% file: tables_2025/maxcut_hyperparams.tex
\begin{table}[htbp]
\centering
\caption{Hyperparameters configurations of the ScoreNet for the \maxcut task.}
\label{tab:cut_config}
\begin{tabular}{lccccc}
\cmidrule[1.5pt]{1-4}
\textbf{Dataset}     & MP units & MP Act & $\delta$ \\
\midrule
G14        & $[32, 32, 32, 32, 16, 16, 16, 16, 8, 8, 8, 8]$ & ReLU & 2.0 \\
G15        & $[32, 32, 32, 32, 16, 16, 16, 16, 8, 8, 8, 8]$ & ReLU & 2.0 \\
G22        & $[4] \times 32$ & TanH & 2.0 \\
G49        & $[32, 32, 32, 32, 16, 16, 16, 16, 8, 8, 8, 8]$ & TanH & 2.0 \\
G50        & $[8] \times 16$ & ReLU & 2.0 \\
G55        & $[4] \times 32$ & ReLU & 2.0 \\
G70        & $[8] \times 16$ & ReLU & 2.0 \\
BarabasiAlbert        & $[4] \times 32$ & TanH & 2.0 \\
Community        & $[4] \times 32$ & TanH & 2.0 \\
ErdősRenyi        & $[4] \times 32$ & TanH & 2.0 \\
Grid2d (10$\times$10)        & $[4] \times 32$ & TanH & 2.0 \\
Grid2d (60$\times$40)        & $[4] \times 32$ & ReLU & 2.0 \\
Minnesota        & $[4] \times 32$ & TanH & 2.0 \\
RandRegular        & $[4] \times 32$ & TanH & 2.0 \\
Ring        & $[4] \times 32$ & ReLU & 2.0 \\
Sensor        & $[4] \times 32$ & TanH & 2.0 \\
\cmidrule[1.5pt]{1-4}
\end{tabular}
\end{table}

%% file: tables_2025/graph_class_hp_merge.tex
\bgroup
\def\arraystretch{0.8} %vertical padding 
\setlength\tabcolsep{.3em} %horizontal padding
\begin{table}[htbp]
\centering
\small
\caption{Hyperparameters configurations of the ScoreNet for the graph classification task.}
\label{tab:graph_class_hp}
\begin{tabular}{l|cc|cc}
\multicolumn{1}{l}{} & \multicolumn{2}{c}{\textbf{\gls{diffndp}}}                               & \multicolumn{2}{c}{\textbf{\gls{diffndp}-E}} \\
\cmidrule[1.5pt]{1-5}
\textbf{Dataset}             & \textbf{\gls{mp} units}                & \boldsymbol{$\beta$} & \textbf{\gls{mp} units}                & \boldsymbol{$\beta$} \\
\midrule
GCB-H        & $[8] \times 16$                                & 3.0     & $[32] \times 8$                                & 5.0     \\
COLLAB       & $[32] \times 8$                                & 1.0     & $[32] \times 8$                                & 1.0     \\
DD           & $[32, 32, 32, 32, 16, 16, 16, 16, 8, 8, 8, 8]$ & 1.0     & $[8] \times 16$                                & 5.0     \\
ENZYMES      & $[8] \times 16$                                & 3.0     & $[16] \times 8$                                & 3.0     \\
EXPWL1       & $[32,32,16,16,8,8]$                            & 1.0     & $[16] \times 8$                                & 1.0     \\
MUTAG        & $[8] \times 16$                                & 1.0     & $[16] \times 8$                                & 3.0     \\
Multipartite & $[32] \times 8$                                & 3.0     & $[32] \times 8$                                & 1.0     \\
Mutagenicity & $[32, 32, 16, 16, 8, 8]$                       & 1.0     & $[32] \times 8$                                & 5.0     \\
NCI1         & $[32, 32, 16, 16, 8, 8]$                       & 1.0     & $[8] \times 16$                                & 3.0     \\
PROTEINS     & $[32, 32, 32, 32, 16, 16, 16, 16, 8, 8, 8, 8]$ & 3.0     & $[32, 32, 16, 16, 8, 8]$                       & 5.0     \\
REDDIT-B     & $[32] \times 8$                                & 1.0     & $[32, 32, 32, 32, 16, 16, 16, 16, 8, 8, 8, 8]$ & 1.0   \\
\cmidrule[1.5pt]{1-5}
\end{tabular}
\end{table}
\egroup

%% file: tables_2025/maxcut_dataset.tex
\bgroup
\def\arraystretch{0.8} %vertical padding 
\setlength\tabcolsep{.3em} %horizontal padding
\begin{table}[htbp]
\centering
\caption{Statistics of the PyGSP datasets used to compute the \maxcut.}
\label{tab:maxcut_dataset}
\subfloat[PyGSP datasets]{
\begin{tabular}{lcccc}
\cmidrule[1.5pt]{1-4}
\textbf{Dataset}     & \textbf{\# Nodes} & \textbf{\# Edges} & \textbf{ Vertex attr.} \\
\midrule
Barabasi-Albert & 100 & 768 & 2 \\
Community & 90 & 532 & 2 \\
Erdős-Renyi & 100 & 974 & 2 \\
Grid2d (10$\times$10) & 100 & 360 & 2 \\
Grid2d (60$\times$40) & 2,400 & 9400 & 2 \\
Minnesota & 2642 & 6608 & 2 \\
RandRegular & 500 & 1500 & 2 \\
Ring & 100 & 200 & 2 \\
Sensor & 64 & 640 & 2 \\
\cmidrule[1.5pt]{1-4}
\end{tabular}
}
~
\subfloat[Gset]{
\begin{tabular}{lcccc}
\cmidrule[1.5pt]{1-4}
\textbf{Dataset}     & \textbf{\# Nodes} & \textbf{\# Edges} & \textbf{ Vertex attr.} \\
\midrule
G14 & $800$ & $4,694$ & -- \\
G15 & $800$ & $4,661$ & -- \\
G22 & $2,000$ & $19,990$ & -- \\
G49 & $3,000$ & $6,000$ & -- \\
G50 & $3,000$ & $6,000$ & -- \\
G55 & $5,000$ & $12,468$ & -- \\
G70 & $10,000$ & $9,999$ & -- \\
\cmidrule[1.5pt]{1-4}
\end{tabular}
}
\end{table}

%% file: algorithms/multipartite_v2.tex
\begin{algorithm}
\caption{Multipartite graph dataset generation}
\label{alg:multipartite}
\small
\begin{algorithmic}[1]
\Require num\_clusters, max\_nodes\_per\_cluster, graphs\_per\_class
\Ensure dataset

\State cluster\_centers $\gets$ GeneratePolygonVertices(num\_clusters) \Comment{Initial arrangement of centers}
\State dataset $\gets \{\}$
\For{class\_label $\gets 0$ to num\_clusters - 1}
    \For{$1$ to graphs\_per\_class}
        \State graph $\gets$ GenerateMultipartiteGraph(cluster\_centers, max\_nodes\_per\_cluster)
        \State graph.label $\gets$ class\_label \Comment{Label based on current rotation}
        \State Add graph to dataset
    \EndFor
    \State cluster\_centers $\gets$ RotateClockwise(cluster\_centers) \Comment{Rotate for next class}
\EndFor
\State \Return dataset
\Statex
\Function{GenerateMultipartiteGraph}{cluster\_centers, max\_nodes\_per\_cluster}
    \For{each center in cluster\_centers}
        \State num\_nodes $\gets$ RandomInt(1, max\_nodes\_per\_cluster)
        \State node\_positions $\gets$ GenerateNodesAroundCenter(center, num\_nodes)
        \State node\_color $\gets$ GetColorForCluster(center) \Comment{Each cluster has a unique color}
        \State AddNodesToGraph(node\_positions, node\_color)
    \EndFor
    \State ConnectNodesAcrossClusters() \Comment{Create complete multipartite graph}
    \State \Return graph
\EndFunction

\Statex
\Function{RotateClockwise}{centers}
    \State \Return [centers[-1]] + centers[:-1] \Comment{Move last center to front}
\EndFunction

\Statex

\end{algorithmic}
\end{algorithm}

%% file: tables_2025/graph_class_stats.tex
% :::::::::::::::::::::::::::: DATASET CLASS ::::::::::::::::::::::::::::
\bgroup
\def\arraystretch{.8} %vertical padding
\setlength\tabcolsep{.5em} %horizontal padding
\begin{table}[!ht]
\caption{\small Details of the graph classification datasets.} 
\label{tab:gc_dataset}
\centering
\begin{tabular}{l|ccccccc}
\cmidrule[1.5pt]{1-8}
\textbf{Dataset} & \textbf{\#Samples} & \textbf{\#Classes} & \textbf{Avg. \#vert.} & \textbf{Avg. \#edg.} & \textbf{V. attr.} & \textbf{V. lab.} & $\boldsymbol{\bar h(\gD)}$ \\
\cmidrule[.5pt]{1-8}
EXPWL1        & 3,000  & 2  & 76.96  & 186.46    & --  & yes & 0.2740 \\
NCI1          & 4,110  & 2  & 29.87  & 64.60     & --  & yes & 0.6245 \\
PROTEINS      & 1,113  & 2  & 39.06  & 72.82     & 1   & yes & 0.6582 \\
Mutagenicity  & 4,337  & 2  & 30.32  & 61.54     & --  & yes & 0.3679 \\
COLLAB        & 5,000  & 3  & 74.49  & 4,914.43  & --  & no  & 1 \\ 
REDDIT-B      & 2,000  & 2  & 429.63 & 497.75    & --  & no  & 1 \\
GCB-H         & 1,800  & 3  & 148.32 & 572.32    & --  & yes & 0.8440 \\
DD            & 1,178  & 2  & 284.32 & 1,431.32  & --  & yes & 0.0688 \\
MUTAG         & 188    & 2  & 17.93  & 19.79     & --  & yes & 0.7082 \\
ENZYMES       & 600    & 6  & 32.63  & 62.14     & 18  & yes & 0.6687 \\
Multipartite  & 5000   & 10 &  99.79 & 4,477.43   & 3   & yes & 0.1101\\
\cmidrule[1.5pt]{1-8}
\end{tabular}
\end{table}

\egroup
% :::::::::::::::::::::::::::::::::::::::::::::::::::::::::::::::::::::

%% file: tables_2025/node_class_stats.tex
\begin{table}[htbp]
\centering
\caption{Statistics of node classification datasets.}
\label{tab:node-class_stats}
\begin{tabular}{lccccc}
\cmidrule[1.5pt]{1-5}
\textbf{Dataset}     & \textbf{\# Nodes} & \textbf{\# Edges} & \textbf{\# Classes} &  $\boldsymbol{h(\gG)}$ \\
\midrule
% Texas                & 183               & 325               & 5                   & \checkmark        & 0.000  \\
% Cornell              & 183               & 298               & 5                   & \checkmark        & 0.150  \\
% Wisconsin            & 521               & 515               & 5                   & \checkmark        & 0.084  \\
% Chameleon            & 2,277             & 31,371            & 5                   & \xmark          & 0.042  \\
% Squirrel             & 5,201             & 198,353           & 5                   & \xmark          & 0.031  \\
Roman-Empire         & 22,662            & 32,927            & 18                             & 0.021  \\
Amazon-Ratings       & 24,492            & 93,050            & 5                            & 0.127  \\
Minesweeper          & 10,000            & 39,402            & 2                              & 0.009  \\
Tolokers             & 11,758            & 519,000           & 2                              & 0.180  \\
Questions            & 48,921            & 153,540           & 2                              & 0.079  \\
\cmidrule[1.5pt]{1-5}
\end{tabular}
\end{table}

%% file: tables_2025/graph_class_res_remaining.tex
\bgroup
\def\arraystretch{.8} %vertical padding
\setlength\tabcolsep{.8em} %horizontal padding
\begin{table}[ht]
\caption{Graph classification accuracy values (subset)}
\label{tab:graph-classification-subset}
\makebox[\textwidth]{
\fontfamily{bch}\selectfont
\centering
\begin{tabular}{@{}l|cccc@{}}
\cmidrule[1.5pt]{1-5}
\textbf{Pooler}  & \textbf{DD} & \textbf{MUTAG} & \textbf{ENZYMES} & \textbf{PROTEINS} \\
\midrule
No pool     & \textcolor{gray}{73{{\tiny$\pm$5}}} & \textcolor{gray}{78{{\tiny$\pm$13}}} & \textcolor{gray}{33{{\tiny$\pm$6}}} & \textcolor{gray}{71{{\tiny$\pm$4}}} \\
\midrule
Diffpool    & 77{{\tiny$\pm$4}} & 81{{\tiny$\pm$11}} & 36{{\tiny$\pm$7}} & 75{{\tiny$\pm$3}} \\
\gls{dmon}        & 78{{\tiny$\pm$5}} & 82{{\tiny$\pm$11}} & 37{{\tiny$\pm$7}} & 76{{\tiny$\pm$4}} \\
\gls{ecpool}    & 73{{\tiny$\pm$5}} & 84{{\tiny$\pm$12}} & 35{{\tiny$\pm$8}} & 74{{\tiny$\pm$5}} \\
Graclus     & 73{{\tiny$\pm$4}} & 82{{\tiny$\pm$12}} & 33{{\tiny$\pm$7}} & 73{{\tiny$\pm$4}} \\
\gls{kmis}        & 75{{\tiny$\pm$3}} & 83{{\tiny$\pm$10}} & 33{{\tiny$\pm$8}} & 73{{\tiny$\pm$5}} \\
\acrlong{mincut}      & 78{{\tiny$\pm$5}} & 81{{\tiny$\pm$12}} & 34{{\tiny$\pm$9}} & 77{{\tiny$\pm$5}} \\
\gls{topk}         & 72{{\tiny$\pm$5}} & 82{{\tiny$\pm$10}} & 29{{\tiny$\pm$7}} & 74{{\tiny$\pm$5}} \\
\midrule
\gls{diffndp}      & 77{{\tiny$\pm$4}} & 84{{\tiny$\pm$10}} & 31{{\tiny$\pm$6}} & 74{{\tiny$\pm$4}} \\
\gls{diffndp}-E   & 77{{\tiny$\pm$3}} & 85{{\tiny$\pm$9}}  & 34{{\tiny$\pm$5}} & 74{{\tiny$\pm$4}} \\
\midrule
\gls{diffndp}-NL  & 74{{\tiny$\pm$4}} & 83{{\tiny$\pm$11}} & 31{{\tiny$\pm$4}} & 70{{\tiny$\pm$4}} \\
\cmidrule[1.5pt]{1-5}
\end{tabular}
}
\end{table}
\egroup

%% file: tables_2025/node_class_cat_res.tex
\bgroup
\def\arraystretch{.8} %vertical padding
\setlength\tabcolsep{.5em} %horizontal padding
\begin{table}[ht]
\caption{Node classification accuracy (Roman-empire, Amazon-ratings) and AUROC (Minesweeper, Tolokers, Questions) obtained when using the architecture with skip connections.}
\label{tab:node-classification-cat}
\makebox[\textwidth]{
\fontfamily{bch}\selectfont
\centering
\begin{tabular}{@{}l|cccccc@{}}
\cmidrule[1.5pt]{1-7}
\textbf{Pooler} & \textbf{Roman-e.} & \textbf{Amazon-r.} & \textbf{Minesw.}$^{*}$ & \textbf{Tolokers} & \textbf{Questions} & \textbf{Score} \\ 
\midrule
\gls{topk}    & 20{\tiny$\pm$11} & 49{\tiny$\pm$7} & 91{{\tiny$\pm$1}} & \textcolor{darkgreen}{96{{\tiny$\pm$0}}} & 70{{\tiny$\pm$3}} & 1 \\
\gls{kmis}    & 19{\tiny$\pm$2}  & \textcolor{darkgreen}{53{\tiny$\pm$3}} & 90{{\tiny$\pm$0}} & 91{{\tiny$\pm$2}} & \textcolor{darkgreen}{82{{\tiny$\pm$4}}} & 2 \\
\gls{ndp}     & 19{\tiny$\pm$4}  & \textcolor{darkgreen}{56{\tiny$\pm$5}} & \textcolor{darkgreen}{94{{\tiny$\pm$0}}} & 90{{\tiny$\pm$8}} & 69{{\tiny$\pm$7}} & 2 \\
\gls{diffndp} & \textcolor{darkgreen}{67{\tiny$\pm$2}}  & 53{\tiny$\pm$1} & 92{{\tiny$\pm$1}} & \textcolor{darkgreen}{96{{\tiny$\pm$1}}} & \textcolor{darkgreen}{82{{\tiny$\pm$2}}} & \textbf{3} \\
\cmidrule[1.5pt]{1-7}
\end{tabular}
}

\end{table}
\egroup

%% file: tables_2025/execution_times.tex
\bgroup
\def\arraystretch{1.0} %vertical padding
\setlength\tabcolsep{.6em} %horizontal padding
\begin{table}[ht]
\caption{Execution times in terms of batches processed per second (b/s) by the architecture for node classification configured with different pooling methods.}
\label{tab:exec_times}
\makebox[\textwidth]{
\fontfamily{bch}\selectfont
\centering
\small
\begin{tabular}{@{}l|ccccc@{}}
\cmidrule[1.5pt]{1-6}
\textbf{Pooler} & \textbf{Roman-e.} & \textbf{Amazon-r.} & \textbf{Minesw.} & \textbf{Tolokers} & \textbf{Questions} \\ 
\midrule
Diffpool          & 0.72 b/s & 0.93 b/s & 0.05 b/s & 0.11 b/s & OOM  \\
\gls{dmon}       & 0.66 b/s & 0.83 b/s & 0.06 b/s & 0.11 b/s & OOM  \\
\acrlong{mincut} & 1.32 b/s & 1.63 b/s & 0.14 b/s & 0.23 b/s & OOM  \\
\gls{topk}       & 0.01 b/s & 0.01 b/s & 0.01 b/s & 0.01 b/s & 0.03 b/s \\
Graclus         & 0.01 b/s & 0.01 b/s & 0.01 b/s & 0.01 b/s & 0.01 b/s \\
\gls{kmis}       & 0.01 b/s & 0.01 b/s & 0.04 b/s & 0.01 b/s & 0.01 b/s \\
\gls{ndp}        & 0.01 b/s & 0.01 b/s & 0.00 b/s & 0.01 b/s & 0.01 b/s \\
\gls{diffndp}    & 0.03 b/s & 0.10 b/s & 0.01 b/s & 0.09 b/s & 0.13 b/s \\
\cmidrule[1.5pt]{1-6}
\end{tabular}
}
\end{table}
\egroup

%% file: tables_2025/memory_usage.tex
\bgroup
\def\arraystretch{1.1} %vertical padding
\setlength\tabcolsep{.4em} %horizontal padding
\begin{table}[ht]
\caption{Average and maximum GPU memory usage (in MB) by the architecture for node classification when configured with different pooling methods.}
\label{tab:mem_usage}
\makebox[\textwidth]{
\fontfamily{bch}\selectfont
\centering
\footnotesize
\begin{tabular}{@{}l|cccccccccc@{}}
\cmidrule[1.5pt]{1-11}
& \multicolumn{2}{l}{\textbf{Roman-e.}} & \multicolumn{2}{l}{\textbf{Amazon-r.}} & \multicolumn{2}{l}{\textbf{Minesw.}} & \multicolumn{2}{l}{\textbf{Tolokers}} & \multicolumn{2}{l}{\textbf{Questions}} \\
\textbf{Pooler}   & \multicolumn{1}{l}{Avg.} & \multicolumn{1}{l}{Max} & \multicolumn{1}{l}{Avg.} & \multicolumn{1}{l}{Max} & \multicolumn{1}{l}{Avg.} & \multicolumn{1}{l}{Max} & \multicolumn{1}{l}{Avg.} & \multicolumn{1}{l}{Max} & \multicolumn{1}{l}{Avg.} & \multicolumn{1}{l}{Max} \\
\hline

Diffpool        & 7167.3 & 11277.2  & 8367.8  & 13165.6   & 1397.4  & 2199.4  & 1931.  & 3039.7  & OOM  & OOM   \\
\gls{dmon}      & 5301.9 & 7359.5   & 6189.5  & 8591.2    & 1035.1  & 1438.2  & 1429.  & 1984.7  & OOM  & OOM   \\
\acrlong{mincut}    & 7167.8 & 11277.6  & 8367.9  & 13165.8   & 1398.0  & 2200.3  & 1932.  & 3040.0  & OOM  & OOM   \\
\gls{topk}      & 2.8    & 3.9      & 3.4     & 4.9       & 1.4     & 1.8     & 3.4    & 6.0     & 5.1  & 8.9 \\
Graclus         & 2.5    & 2.6      & 4.0     & 4.1       & 1.5     & 1.5     & 14.6   & 15.1    & 10.4 & 10.6  \\
\gls{kmis}      & 1.3    & 1.3      & 0.7     & 0.8       & 0.3     & 0.3     & 2.4    & 2.5     & 5.2  & 5.3   \\
\gls{ndp}       & 1.8    & 2.5      & 1.8     & 9.7       & 0.6     & 2.5     & 2.4    & 70.8    & 2.4  & 26.1  \\
\gls{diffndp}   & 13.7   & 25.7     & 16.8    & 31.2      & 6.9     & 12.7    & 27.9   & 52.2    & 32.6 & 61.6  \\
\cmidrule[1.5pt]{1-11}
\end{tabular}
}
\end{table}
\egroup

% \bgroup
% \def\arraystretch{.8} %vertical padding
% \setlength\tabcolsep{.5em} %horizontal padding
% \begin{table}[ht]
% \caption{Execution times in terms of batches processed per second by the architecture for node classification configured with different pooling methods.}
% \label{tab:exec_times}
% \makebox[\textwidth]{
% \fontfamily{bch}\selectfont
% \centering
% \begin{tabular}{@{}l|ccccc@{}}
% \cmidrule[1.5pt]{1-6}
% \textbf{Pooler} & \textbf{Roman-e.} & \textbf{Amazon-r.} & \textbf{Minesw.} & \textbf{Tolokers} & \textbf{Questions} \\ 
% \midrule
% Diffpool          & 0.72 b/s & 0.93 b/s & 0.05 b/s & 0.11 b/s & OOM  \\
% \gls{dmon}       & 0.66 b/s & 0.83 b/s & 0.06 b/s & 0.11 b/s & OOM  \\
% \acrlong{mincut} & 1.32 b/s & 1.63 b/s & 0.14 b/s & 0.23 b/s & OOM  \\
% \gls{topk}       & 0.01 b/s & 0.01 b/s & 0.01 b/s & 0.01 b/s & 0.03 b/s \\
% Graclus         & 0.01 b/s & 0.01 b/s & 0.01 b/s & 0.01 b/s & 0.01 b/s \\
% \gls{kmis}       & 0.01 b/s & 0.01 b/s & 0.04 b/s & 0.01 b/s & 0.01 b/s \\
% \gls{ndp}        & 0.01 b/s & 0.01 b/s & 0.00 b/s & 0.01 b/s & 0.01 b/s \\
% \gls{diffndp}    & 0.03 b/s & 0.10 b/s & 0.01 b/s & 0.09 b/s & 0.13 b/s \\
% \cmidrule[1.5pt]{1-6}
% \end{tabular}
% }
% \end{table}
% \egroup

%% file: main.bbl
\begin{thebibliography}{72}
\providecommand{\natexlab}[1]{#1}
\providecommand{\url}[1]{\texttt{#1}}
\expandafter\ifx\csname urlstyle\endcsname\relax
  \providecommand{\doi}[1]{doi: #1}\else
  \providecommand{\doi}{doi: \begingroup \urlstyle{rm}\Url}\fi

\bibitem[Abbas \& Swoboda(2022)Abbas and Swoboda]{abbas2022rama}
Ahmed Abbas and Paul Swoboda.
\newblock Rama: A rapid multicut algorithm on gpu.
\newblock In \emph{Proceedings of the IEEE/CVF Conference on Computer Vision and Pattern Recognition}, pp.\  8193--8202, 2022.

\bibitem[Aspvall \& Gilbert(1984)Aspvall and Gilbert]{1984Graph}
Bengt Aspvall and John~R. Gilbert.
\newblock Graph coloring using eigenvalue decomposition.
\newblock \emph{SIAM Journal on Algebraic Discrete Methods}, 5\penalty0 (4):\penalty0 526–538, 1984.

\bibitem[Bacciu et~al.(2023)Bacciu, Conte, and Landolfi]{bacciu2023pooling}
Davide Bacciu, Alessio Conte, and Francesco Landolfi.
\newblock Graph pooling with maximum-weight $k$-independent sets.
\newblock In \emph{Thirty-Seventh AAAI Conference on Artificial Intelligence}, 2023.

\bibitem[Bashar et~al.(2020)Bashar, Mallick, Truesdell, Calhoun, Joshi, and Shukla]{9204635}
Mohammad~Khairul Bashar, Antik Mallick, Daniel~S. Truesdell, Benton~H. Calhoun, Siddharth Joshi, and Nikhil Shukla.
\newblock Experimental demonstration of a reconfigurable coupled oscillator platform to solve the max-cut problem.
\newblock \emph{IEEE Journal on Exploratory Solid-State Computational Devices and Circuits}, 6\penalty0 (2):\penalty0 116--121, 2020.
\newblock \doi{10.1109/JXCDC.2020.3025994}.

\bibitem[Bianchi(2022)]{bianchi2022simplifying}
Filippo~Maria Bianchi.
\newblock Simplifying clustering with graph neural networks.
\newblock \emph{arXiv preprint arXiv:2207.08779}, 2022.

\bibitem[Bianchi \& Lachi(2023)Bianchi and Lachi]{bianchi2023expr}
Filippo~Maria Bianchi and Veronica Lachi.
\newblock The expressive power of pooling in graph neural networks.
\newblock In \emph{Advances in Neural Information Processing Systems}, volume~36, pp.\  71603--71618, 2023.

\bibitem[Bianchi et~al.(2020{\natexlab{a}})Bianchi, Grattarola, and Alippi]{bianchi2020spectral}
Filippo~Maria Bianchi, Daniele Grattarola, and Cesare Alippi.
\newblock Spectral clustering with graph neural networks for graph pooling.
\newblock In \emph{International conference on machine learning}, pp.\  874--883. PMLR, 2020{\natexlab{a}}.

\bibitem[Bianchi et~al.(2020{\natexlab{b}})Bianchi, Grattarola, Livi, and Alippi]{bianchi2020hierarchical}
Filippo~Maria Bianchi, Daniele Grattarola, Lorenzo Livi, and Cesare Alippi.
\newblock Hierarchical representation learning in graph neural networks with node decimation pooling.
\newblock \emph{IEEE Transactions on Neural Networks and Learning Systems}, 33\penalty0 (5):\penalty0 2195--2207, 2020{\natexlab{b}}.

\bibitem[Bianchi et~al.(2021)Bianchi, Grattarola, Livi, and Alippi]{bianchi2021graph}
Filippo~Maria Bianchi, Daniele Grattarola, Lorenzo Livi, and Cesare Alippi.
\newblock Graph neural networks with convolutional arma filters.
\newblock \emph{IEEE transactions on pattern analysis and machine intelligence}, 44\penalty0 (7):\penalty0 3496--3507, 2021.

\bibitem[Bianchi et~al.(2022)Bianchi, Gallicchio, and Micheli]{bianchi2022pyramidal}
Filippo~Maria Bianchi, Claudio Gallicchio, and Alessio Micheli.
\newblock Pyramidal reservoir graph neural network.
\newblock \emph{Neurocomputing}, 470:\penalty0 389--404, 2022.
\newblock ISSN 0925-2312.
\newblock \doi{https://doi.org/10.1016/j.neucom.2021.04.131}.

\bibitem[Blakely et~al.(2021)Blakely, Lanchantin, and Qi]{blakely2019complexity}
Derrick Blakely, Jack Lanchantin, and Yanjun Qi.
\newblock Time and space complexity of graph convolutional networks.
\newblock \emph{Accessed on: Dec}, 31:\penalty0 2021, 2021.

\bibitem[Bommasani et~al.(2021)Bommasani, Hudson, Adeli, Altman, Arora, von Arx, Bernstein, Bohg, Bosselut, Brunskill, et~al.]{bommasani2021opportunities}
Rishi Bommasani, Drew~A Hudson, Ehsan Adeli, Russ Altman, Simran Arora, Sydney von Arx, Michael~S Bernstein, Jeannette Bohg, Antoine Bosselut, Emma Brunskill, et~al.
\newblock On the opportunities and risks of foundation models.
\newblock \emph{arXiv preprint arXiv:2108.07258}, 2021.

\bibitem[Borgs et~al.(2012)Borgs, Chayes, Lov{\'a}sz, S{\'o}s, and Vesztergombi]{borgs2012convergent}
Christian Borgs, Jennifer~T Chayes, L{\'a}szl{\'o} Lov{\'a}sz, Vera~T S{\'o}s, and Katalin Vesztergombi.
\newblock Convergent sequences of dense graphs ii. multiway cuts and statistical physics.
\newblock \emph{Annals of Mathematics}, pp.\  151--219, 2012.

\bibitem[Chien et~al.(2021)Chien, Peng, Li, and Milenkovic]{chien2021adaptive}
Eli Chien, Jianhao Peng, Pan Li, and Olgica Milenkovic.
\newblock Adaptive universal generalized pagerank graph neural network.
\newblock In \emph{International Conference on Learning Representations}, 2021.

\bibitem[Cini et~al.(2024)Cini, Mandic, and Alippi]{cini2024graph}
Andrea Cini, Danilo Mandic, and Cesare Alippi.
\newblock {Graph-based Time Series Clustering for End-to-End Hierarchical Forecasting}.
\newblock \emph{International Conference on Machine Learning}, 2024.

\bibitem[Defferrard et~al.(2016)Defferrard, Bresson, and Vandergheynst]{defferrard2016convolutional}
Micha{\"e}l Defferrard, Xavier Bresson, and Pierre Vandergheynst.
\newblock Convolutional neural networks on graphs with fast localized spectral filtering.
\newblock \emph{Advances in neural information processing systems}, 29, 2016.

\bibitem[Defferrard et~al.(2017)Defferrard, Martin, Pena, and Perraudin]{defferrard2017pygsp}
Michaël Defferrard, Lionel Martin, Rodrigo Pena, and Nathanaël Perraudin.
\newblock Pygsp: Graph signal processing in python, October 2017.
\newblock URL \url{https://doi.org/10.5281/zenodo.1003158}.

\bibitem[Dhillon et~al.(2007)Dhillon, Guan, and Kulis]{dhillon2007weighted}
Inderjit~S Dhillon, Yuqiang Guan, and Brian Kulis.
\newblock Weighted graph cuts without eigenvectors a multilevel approach.
\newblock \emph{IEEE transactions on pattern analysis and machine intelligence}, 29\penalty0 (11):\penalty0 1944--1957, 2007.

\bibitem[Diehl(2019)]{diehl2019edge}
Frederik Diehl.
\newblock Edge contraction pooling for graph neural networks.
\newblock \emph{arXiv preprint arXiv:1905.10990}, 2019.

\bibitem[Dong et~al.(2021)Dong, Ding, Jalaian, Ji, and Li]{dong2021adagnn}
Yushun Dong, Kaize Ding, Brian Jalaian, Shuiwang Ji, and Jundong Li.
\newblock Adagnn: Graph neural networks with adaptive frequency response filter.
\newblock In \emph{Proceedings of the 30th ACM international conference on information \& knowledge management}, pp.\  392--401, 2021.

\bibitem[Duval \& Malliaros(2022)Duval and Malliaros]{duval2022higher}
Alexandre Duval and Fragkiskos Malliaros.
\newblock Higher-order clustering and pooling for graph neural networks.
\newblock In \emph{Proceedings of the 31st ACM international conference on information \& knowledge management}, pp.\  426--435, 2022.

\bibitem[Eliasof et~al.(2023)Eliasof, Ruthotto, and Treister]{eliasof2023improving}
Moshe Eliasof, Lars Ruthotto, and Eran Treister.
\newblock Improving graph neural networks with learnable propagation operators.
\newblock In \emph{International Conference on Machine Learning}, pp.\  9224--9245. PMLR, 2023.

\bibitem[Errica et~al.(2020)Errica, Podda, Bacciu, and Micheli]{Errica2020A}
Federico Errica, Marco Podda, Davide Bacciu, and Alessio Micheli.
\newblock A fair comparison of graph neural networks for graph classification.
\newblock In \emph{International Conference on Learning Representations}, 2020.

\bibitem[Fey \& Lenssen(2019)Fey and Lenssen]{fey2019fast}
Matthias Fey and Jan~E. Lenssen.
\newblock Fast graph representation learning with {PyTorch Geometric}.
\newblock In \emph{ICLR Workshop on Representation Learning on Graphs and Manifolds}, 2019.

\bibitem[Fu et~al.(2022)Fu, Zhao, and Bian]{fu2022p}
Guoji Fu, Peilin Zhao, and Yatao Bian.
\newblock $ p $-laplacian based graph neural networks.
\newblock In \emph{International Conference on Machine Learning}, pp.\  6878--6917. PMLR, 2022.

\bibitem[Gao et~al.(2021)Gao, Liu, and Ji]{gao2021topologytapool}
H.~Gao, Y.~Liu, and S.~Ji.
\newblock Topology-aware graph pooling networks.
\newblock \emph{IEEE Transactions on Pattern Analysis and Machine Intelligence}, 43\penalty0 (12):\penalty0 4512--4518, dec 2021.
\newblock ISSN 1939-3539.
\newblock \doi{10.1109/TPAMI.2021.3062794}.

\bibitem[Gao \& Ji(2019)Gao and Ji]{gao2019graph}
Hongyang Gao and Shuiwang Ji.
\newblock Graph u-nets.
\newblock In \emph{international conference on machine learning}, pp.\  2083--2092. PMLR, 2019.

\bibitem[Gao et~al.(2022)Gao, Dai, Li, Xiong, and Frossard]{gao2022ipool}
Xing Gao, Wenrui Dai, Chenglin Li, Hongkai Xiong, and Pascal Frossard.
\newblock ipool—information-based pooling in hierarchical graph neural networks.
\newblock \emph{IEEE Transactions on Neural Networks and Learning Systems}, 33\penalty0 (9):\penalty0 5032--5044, 2022.
\newblock \doi{10.1109/TNNLS.2021.3067441}.

\bibitem[Goemans \& Williamson(1995)Goemans and Williamson]{goemanswilliamson1995improved}
Michel~X. Goemans and David~P. Williamson.
\newblock Improved approximation algorithms for maximum cut and satisfiability problems using semidefinite programming.
\newblock \emph{J. ACM}, 42\penalty0 (6):\penalty0 1115–1145, nov 1995.
\newblock ISSN 0004-5411.
\newblock \doi{10.1145/227683.227684}.

\bibitem[Grattarola et~al.(2022)Grattarola, Zambon, Bianchi, and Alippi]{grattarola2022understanding}
Daniele Grattarola, Daniele Zambon, Filippo~Maria Bianchi, and Cesare Alippi.
\newblock Understanding pooling in graph neural networks.
\newblock \emph{IEEE Transactions on Neural Networks and Learning Systems}, 2022.

\bibitem[Hansen \& Bianchi(2023)Hansen and Bianchi]{hansen2023total}
Jonas~Berg Hansen and Filippo~Maria Bianchi.
\newblock Total variation graph neural networks.
\newblock In \emph{International Conference on Machine Learning}, pp.\  12445--12468. PMLR, 2023.

\bibitem[Hu et~al.(2020)Hu, Liu, Gomes, Zitnik, Liang, Pande, and Leskovec]{hu2020strategies-gine}
Weihua Hu, Bowen Liu, Joseph Gomes, Marinka Zitnik, Percy Liang, Vijay Pande, and Jure Leskovec.
\newblock Strategies for pre-training graph neural networks, 2020.

\bibitem[Jin et~al.(2020)Jin, Loukas, and JaJa]{jin2020graph}
Yu~Jin, Andreas Loukas, and Joseph JaJa.
\newblock Graph coarsening with preserved spectral properties.
\newblock In \emph{International Conference on Artificial Intelligence and Statistics}, pp.\  4452--4462. PMLR, 2020.

\bibitem[Khasahmadi et~al.(2020)Khasahmadi, Hassani, Moradi, Lee, and Morris]{Khasahmadi2020Memory-Based}
Amir~Hosein Khasahmadi, Kaveh Hassani, Parsa Moradi, Leo Lee, and Quaid Morris.
\newblock Memory-based graph networks.
\newblock In \emph{International Conference on Learning Representations}, 2020.

\bibitem[Kingma \& Ba(2015)Kingma and Ba]{DBLP:journals/corr/KingmaB14}
Diederik~P. Kingma and Jimmy Ba.
\newblock Adam: {A} method for stochastic optimization.
\newblock In \emph{3rd International Conference on Learning Representations, {ICLR} 2015, San Diego, CA, USA, May 7-9, 2015, Conference Track Proceedings}, 2015.

\bibitem[Kipf \& Welling(2017)Kipf and Welling]{kipf2017semisupervised}
Thomas~N. Kipf and Max Welling.
\newblock Semi-supervised classification with graph convolutional networks.
\newblock In \emph{5th International Conference on Learning Representations, {ICLR} 2017, Toulon, France, April 24-26, 2017, Conference Track Proceedings}. OpenReview.net, 2017.

\bibitem[Knyazev et~al.(2019)Knyazev, Taylor, and Amer]{knyazev2019understanding}
Boris Knyazev, Graham~W Taylor, and Mohamed Amer.
\newblock Understanding attention and generalization in graph neural networks.
\newblock \emph{Advances in neural information processing systems}, 32, 2019.

\bibitem[Landolfi(2022)]{landolfi2022revisiting}
Francesco Landolfi.
\newblock Revisiting edge pooling in graph neural networks.
\newblock In \emph{ESANN}, 2022.

\bibitem[Lee et~al.(2019)Lee, Lee, and Kang]{lee2019self}
Junhyun Lee, Inyeop Lee, and Jaewoo Kang.
\newblock Self-attention graph pooling.
\newblock In \emph{International conference on machine learning}, pp.\  3734--3743. PMLR, 2019.

\bibitem[Lee \& Constantinides(2023)Lee and Constantinides]{lee2023quantumized}
W~Bernard Lee and Anthony~G Constantinides.
\newblock Quantumized graph cuts in portfolio construction and asset selection.
\newblock \emph{Springer-Nature Transactions on Computational Science and Compu-tational Intelligence}, 2023.

\bibitem[Liers et~al.(2004)Liers, J{\"u}nger, Reinelt, and Rinaldi]{liers2004computing}
Frauke Liers, Michael J{\"u}nger, Gerhard Reinelt, and Giovanni Rinaldi.
\newblock Computing exact ground states of hard ising spin glass problems by branch-and-cut.
\newblock \emph{New optimization algorithms in physics}, pp.\  47--69, 2004.

\bibitem[Lim et~al.(2021)Lim, Hohne, Li, Huang, Gupta, Bhalerao, and Lim]{lim2021large}
Derek Lim, Felix Hohne, Xiuyu Li, Sijia~Linda Huang, Vaishnavi Gupta, Omkar Bhalerao, and Ser~Nam Lim.
\newblock Large scale learning on non-homophilous graphs: New benchmarks and strong simple methods.
\newblock \emph{Advances in Neural Information Processing Systems}, 34:\penalty0 20887--20902, 2021.

\bibitem[Liu et~al.(2021)Liu, Jian, Li, Zhang, Lai, and Xu]{liu2021hierarchical}
Ning Liu, Songlei Jian, Dongsheng Li, Yiming Zhang, Zhiquan Lai, and Hongzuo Xu.
\newblock Hierarchical adaptive pooling by capturing high-order dependency for graph representation learning.
\newblock \emph{IEEE Transactions on Knowledge and Data Engineering}, 35\penalty0 (4):\penalty0 3952--3965, 2021.

\bibitem[Loukas(2019)]{loukas2019graph}
Andreas Loukas.
\newblock Graph reduction with spectral and cut guarantees.
\newblock \emph{Journal of Machine Learning Research}, 20\penalty0 (116):\penalty0 1--42, 2019.

\bibitem[Luzhnica et~al.(2019)Luzhnica, Day, and Lio]{luzhnica2019clique}
Enxhell Luzhnica, Ben Day, and Pietro Lio.
\newblock Clique pooling for graph classification.
\newblock \emph{arXiv preprint arXiv:1904.00374}, 2019.

\bibitem[Ma et~al.(2020)Ma, Xuan, Wang, Li, and Li{\`o}]{ma2020path}
Zheng Ma, Junyu Xuan, Yu~Guang Wang, Ming Li, and Pietro Li{\`o}.
\newblock Path integral based convolution and pooling for graph neural networks.
\newblock \emph{Advances in Neural Information Processing Systems}, 33:\penalty0 16421--16433, 2020.

\bibitem[Makarychev et~al.(2014)Makarychev, Makarychev, and Vijayaraghavan]{makarychev2014bilu}
Konstantin Makarychev, Yury Makarychev, and Aravindan Vijayaraghavan.
\newblock Bilu--linial stable instances of max cut and minimum multiway cut.
\newblock In \emph{Proceedings of the twenty-fifth annual ACM-SIAM symposium on Discrete algorithms}, pp.\  890--906. SIAM, 2014.

\bibitem[Marisca et~al.(2024)Marisca, Alippi, and Bianchi]{marisca2024graph}
Ivan Marisca, Cesare Alippi, and Filippo~Maria Bianchi.
\newblock Graph-based forecasting with missing data through spatiotemporal downsampling.
\newblock In \emph{Proceedings of the 41st International Conference on Machine Learning}, volume 235 of \emph{Proceedings of Machine Learning Research}, pp.\  34846--34865. PMLR, 2024.

\bibitem[Mather \& Koch(2022)Mather and Koch]{mather2022computer}
Paul~M Mather and Magaly Koch.
\newblock \emph{Computer processing of remotely-sensed images}.
\newblock John Wiley \& Sons, 2022.

\bibitem[Morris et~al.(2020)Morris, Kriege, Bause, Kersting, Mutzel, and Neumann]{morris2020tudataset}
Christopher Morris, Nils~M. Kriege, Franka Bause, Kristian Kersting, Petra Mutzel, and Marion Neumann.
\newblock Tudataset: A collection of benchmark datasets for learning with graphs.
\newblock In \emph{ICML 2020 Workshop on Graph Representation Learning and Beyond (GRL+ 2020)}, 2020.

\bibitem[Noutahi et~al.(2019)Noutahi, Beaini, Horwood, Gigu{\`e}re, and Tossou]{noutahi2019towards}
Emmanuel Noutahi, Dominique Beaini, Julien Horwood, S{\'e}bastien Gigu{\`e}re, and Prudencio Tossou.
\newblock Towards interpretable sparse graph representation learning with laplacian pooling.
\newblock \emph{arXiv preprint arXiv:1905.11577}, 2019.

\bibitem[Pang et~al.(2021)Pang, Zhao, and Li]{pang2021graph}
Yunsheng Pang, Yunxiang Zhao, and Dongsheng Li.
\newblock Graph pooling via coarsened graph infomax.
\newblock In \emph{Proceedings of the 44th International ACM SIGIR Conference on Research and Development in Information Retrieval}, pp.\  2177--2181, 2021.

\bibitem[Platonov et~al.(2023)Platonov, Kuznedelev, Diskin, Babenko, and Prokhorenkova]{platonov2023a}
Oleg Platonov, Denis Kuznedelev, Michael Diskin, Artem Babenko, and Liudmila Prokhorenkova.
\newblock A critical look at the evaluation of {GNN}s under heterophily: Are we really making progress?
\newblock In \emph{The Eleventh International Conference on Learning Representations}, 2023.

\bibitem[Ranjan et~al.(2020)Ranjan, Sanyal, and Talukdar]{ranjan2020asap}
Ekagra Ranjan, Soumya Sanyal, and Partha Talukdar.
\newblock Asap: Adaptive structure aware pooling for learning hierarchical graph representations.
\newblock In \emph{Proceedings of the AAAI Conference on Artificial Intelligence}, volume~34, pp.\  5470--5477, 2020.

\bibitem[Schuetz et~al.(2022)Schuetz, Brubaker, and Katzgraber]{schuetz2022combinatorial}
Martin~JA Schuetz, J~Kyle Brubaker, and Helmut~G Katzgraber.
\newblock Combinatorial optimization with physics-inspired graph neural networks.
\newblock \emph{Nature Machine Intelligence}, 4\penalty0 (4):\penalty0 367--377, 2022.

\bibitem[Shuman et~al.(2013)Shuman, Narang, Frossard, Ortega, and Vandergheynst]{shuman2013emerging}
David~I Shuman, Sunil~K. Narang, Pascal Frossard, Antonio Ortega, and Pierre Vandergheynst.
\newblock The emerging field of signal processing on graphs: Extending high-dimensional data analysis to networks and other irregular domains.
\newblock \emph{IEEE Signal Processing Magazine}, 30\penalty0 (3):\penalty0 83--98, 2013.
\newblock \doi{10.1109/MSP.2012.2235192}.

\bibitem[Shuman et~al.(2015)Shuman, Faraji, and Vandergheynst]{shuman2015multiscale}
David~I Shuman, Mohammad~Javad Faraji, and Pierre Vandergheynst.
\newblock A multiscale pyramid transform for graph signals.
\newblock \emph{IEEE Transactions on Signal Processing}, 64\penalty0 (8):\penalty0 2119--2134, 2015.

\bibitem[Tremblay et~al.(2018)Tremblay, Gon{\c{c}}alves, and Borgnat]{tremblay2018design}
Nicolas Tremblay, Paulo Gon{\c{c}}alves, and Pierre Borgnat.
\newblock Design of graph filters and filterbanks.
\newblock In \emph{Cooperative and Graph Signal Processing}, pp.\  299--324. Elsevier, 2018.

\bibitem[Trevisan(2009)]{trevisan2009max}
Luca Trevisan.
\newblock Max cut and the smallest eigenvalue.
\newblock In \emph{Proceedings of the forty-first annual ACM symposium on Theory of computing}, pp.\  263--272, 2009.

\bibitem[Tsitsulin et~al.(2023)Tsitsulin, Palowitch, Perozzi, and M{\"{u}}ller]{tsitsulin2020graph}
Anton Tsitsulin, John Palowitch, Bryan Perozzi, and Emmanuel M{\"{u}}ller.
\newblock Graph clustering with graph neural networks.
\newblock \emph{J. Mach. Learn. Res.}, 24:\penalty0 127:1--127:21, 2023.

\bibitem[von Luxburg(2007)]{vonluxburg2007tutorial}
Ulrike von Luxburg.
\newblock A tutorial on spectral clustering, 2007.

\bibitem[Wang et~al.(2019)Wang, Ying, Huang, and Leskovec]{wang2019oversmoothing}
Guangtao Wang, Rex Ying, Jing Huang, and Jure Leskovec.
\newblock Improving graph attention networks with large margin-based constraints, 2019.

\bibitem[Wang et~al.(2024)Wang, Luo, Shen, Heng, and Luo]{wang2024comprehensive}
Pengyun Wang, Junyu Luo, Yanxin Shen, Siyu Heng, and Xiao Luo.
\newblock A comprehensive graph pooling benchmark: Effectiveness, robustness and generalizability.
\newblock \emph{arXiv preprint arXiv:2406.09031}, 2024.

\bibitem[Wu et~al.(2019)Wu, Souza, Zhang, Fifty, Yu, and Weinberger]{wu2019oversmoothing}
Felix Wu, Amauri Souza, Tianyi Zhang, Christopher Fifty, Tao Yu, and Kilian Weinberger.
\newblock Simplifying graph convolutional networks.
\newblock In Kamalika Chaudhuri and Ruslan Salakhutdinov (eds.), \emph{Proceedings of the 36th International Conference on Machine Learning}, volume~97 of \emph{Proceedings of Machine Learning Research}, pp.\  6861--6871. PMLR, 09--15 Jun 2019.

\bibitem[Wu et~al.(2022)Wu, Chen, Xu, and Li]{wu2022structural}
Junran Wu, Xueyuan Chen, Ke~Xu, and Shangzhe Li.
\newblock Structural entropy guided graph hierarchical pooling.
\newblock In \emph{International conference on machine learning}, pp.\  24017--24030. PMLR, 2022.

\bibitem[Xu et~al.(2019)Xu, Hu, Leskovec, and Jegelka]{xu2018powerful}
Keyulu Xu, Weihua Hu, Jure Leskovec, and Stefanie Jegelka.
\newblock How powerful are graph neural networks?
\newblock In \emph{International Conference on Learning Representations}, 2019.

\bibitem[Ye(2003)]{ye2003gset}
Yinyu Ye.
\newblock The gset dataset, 2003.

\bibitem[Ying et~al.(2018)Ying, You, Morris, Ren, Hamilton, and Leskovec]{ying2018hierarchical}
Zhitao Ying, Jiaxuan You, Christopher Morris, Xiang Ren, Will Hamilton, and Jure Leskovec.
\newblock Hierarchical graph representation learning with differentiable pooling.
\newblock \emph{Advances in neural information processing systems}, 31, 2018.

\bibitem[Yuan \& Ji(2020)Yuan and Ji]{yuan2020structpool}
Hao Yuan and Shuiwang Ji.
\newblock Structpool: Structured graph pooling via conditional random fields.
\newblock In \emph{Proceedings of the 8th International Conference on Learning Representations}, 2020.

\bibitem[Zhang et~al.(2019)Zhang, Bu, Ester, Zhang, Yao, Yu, and Wang]{zhang2019hierarchical}
Zhen Zhang, Jiajun Bu, Martin Ester, Jianfeng Zhang, Chengwei Yao, Zhi Yu, and Can Wang.
\newblock Hierarchical graph pooling with structure learning.
\newblock \emph{arXiv preprint arXiv:1911.05954}, 2019.

\bibitem[Zhou et~al.(2020{\natexlab{a}})Zhou, Cui, Hu, Zhang, Yang, Liu, Wang, Li, and Sun]{zhou2020review}
Jie Zhou, Ganqu Cui, Shengding Hu, Zhengyan Zhang, Cheng Yang, Zhiyuan Liu, Lifeng Wang, Changcheng Li, and Maosong Sun.
\newblock Graph neural networks: A review of methods and applications.
\newblock \emph{AI Open}, 1:\penalty0 57--81, 2020{\natexlab{a}}.
\newblock ISSN 2666-6510.
\newblock \doi{10.1016/j.aiopen.2021.01.001}.

\bibitem[Zhou et~al.(2020{\natexlab{b}})Zhou, Wang, Choi, Pichler, and Lukin]{zhou2020quantum}
Leo Zhou, Sheng-Tao Wang, Soonwon Choi, Hannes Pichler, and Mikhail~D Lukin.
\newblock Quantum approximate optimization algorithm: Performance, mechanism, and implementation on near-term devices.
\newblock \emph{Physical Review X}, 10\penalty0 (2):\penalty0 021067, 2020{\natexlab{b}}.

\end{thebibliography}
